\documentclass[sigconf]{acmart}

\usepackage{stfloats}

\copyrightyear{2026}
\acmYear{2026}
\setcopyright{cc}
\setcctype{by}
\acmConference[SIGSPATIAL '26]{The 34th ACM International Conference on Advances in Geographic Information Systems}{November 03--06, 2026}{Riverside, CA, USA}
\acmBooktitle{The 34th ACM International Conference on Advances in Geographic Information Systems (SIGSPATIAL '26), November 03--06, 2026, Riverside, CA, USA}
\acmDOI{10.1145/3841645.3843351}
\acmISBN{979-8-4007-2950-8/2026/11}

\begin{document}

\title[Double-Diffusion]{Double-Diffusion: Balancing Speed, Accuracy, and Uncertainty in Probabilistic Forecasting for Urban Sensor Networks}

\author{Hanlin Dong}
\email{edward.dong1@unsw.edu.au}
\affiliation{\institution{University of New South Wales}\city{Sydney}\country{Australia}}
\author{Arian Prabowo}
\email{arian.prabowo@unsw.edu.au}
\affiliation{\institution{University of New South Wales}\city{Sydney}\country{Australia}}
\author{Hao Xue}
\email{hao.xue1@unsw.edu.au}
\affiliation{\institution{University of New South Wales}\city{Sydney}\country{Australia}}
\affiliation{\institution{The Hong Kong University of Science and Technology (Guangzhou)}\city{Guangzhou}\country{China}}
\author{Shuang Ao}
\email{shuang.ao@unsw.edu.au}
\affiliation{\institution{Zhejiang University of Technology}\city{Hangzhou}\country{China}}
\author{Tianyi Zhou}
\email{tianyi.david.zhou@gmail.com}
\affiliation{\institution{Mohamed bin Zayed University of Artificial Intelligence}\city{Abu Dhabi}\country{United Arab Emirates}}
\author{Yuxuan Liang}
\email{yuxliang@outlook.com}
\affiliation{\institution{The Hong Kong
University of Science and Technology
(Guangzhou)
}\city{Guangzhou}\country{China}}
\author{Flora D. Salim}
\email{flora.salim@unsw.edu.au}
\affiliation{\institution{University of New South Wales}\city{Sydney}\country{Australia}}
\begin{abstract}
Urban sensor networks need forecasts that are accurate, carry useful uncertainty, and refresh fast enough to act on as new readings arrive. These goals conflict: deterministic models give no distribution, while diffusion forecasters model uncertainty but denoise from pure noise over many steps. We present Double-Diffusion, which integrates a closed-form graph-heat prior into a denoising diffusion model. The prior is a parameter-free low-pass forecast over the sensor graph. It serves two roles: it shifts the forward process the model learns to invert, and it conditions the denoiser. The reverse process therefore starts from the noised prior and recovers the forecast over a truncated chain instead of synthesizing from pure noise; the name denotes this composition, a graph diffusion feeding a denoising diffusion. A compact denoiser, DD-Net, is trained as a Denoising Diffusion Probabilistic Model (DDPM) under this residual-shifted forward process, and an offline graph-spectral gate diagnostic reads the low-band energy of the prior residual to set the denoiser's switchable spatial filter per domain from training data alone. On four real-world air quality and traffic networks, Double-Diffusion matches or beats every evaluated probabilistic forecaster on CRPS: it attains the lowest observed mean CRPS on three of the four networks, a ranking that holds identically in the initial and rolling views, and ranks second within 3\% of the best on the fourth, while staying competitive in point accuracy with the strongest baselines. Because generation begins partway along the schedule, sampling runs only a truncated chain: measured across the four networks, Double-Diffusion samples faster than the evaluated full-chain diffusion forecasters at their own operating points, and its sampling cost grows sublinearly with the scheduled chain length where full-chain sampling grows linearly, so the margin widens as schedules lengthen. The code is available at: https://github.com/teddyicare/Double-Diffusion
\end{abstract}

\ccsdesc[500]{Computing methodologies~Machine learning}
\ccsdesc[300]{Information systems~Geographic information systems}

\keywords{spatio-temporal forecasting, diffusion models, uncertainty quantification, urban computing, graph neural networks}

\maketitle

\section{Introduction}

From air quality monitors to freeway loop detectors, urban sensor networks produce continuous graph-structured time series that underpin monitoring, alerting, and operational decision making. Operators want all three of accuracy, meaningful uncertainty for risk decisions, and inference fast enough to repeat as new observations arrive, and the three conflict. Deterministic models such as GMAN~\cite{zheng_gman_2020} and GMSDR~\cite{liu_msdr_2022} are fast and accurate but emit no distribution. Probabilistic forecasters such as DeepAR~\cite{salinas_deepar_2020} and MC Dropout~\cite{gal_dropout_2016} model uncertainty but tend to be less accurate~\cite{wen_diffstg_2023}. Diffusion forecasters such as CSDI~\cite{tashiro_csdi_2021} and DiffSTG~\cite{wen_diffstg_2023} give expressive distributions and better point accuracy, but run a long reverse process from pure noise over tens to hundreds of denoising steps, which makes them substantially slower at inference.

While a diffusion forecaster that starts from pure Gaussian noise eventually recovers the field, it spends most of its steps rebuilding coarse, predictable structure, the spatial smoothness across nearby sensors, before it refines any detail. That coarse structure is exactly what a cheap, closed-form prior can supply directly. We present Double-Diffusion (DD), which replaces the noise start with such a prior. A closed-form graph diffusion propagates the most recent observation over the sensor graph, with no learned parameters, into a spatially smooth preliminary forecast. Rather than train a separate corrector, we adapt the residual-shift idea from diffusion-based image restoration~\cite{shi_resfusion_2024, liu_residual_2024}, where the reverse process starts from a degraded input rather than pure noise. Forecasting supplies no such input, so we synthesize one: the prior plays the part the blurred photograph plays in restoration, a rough version of the answer that the model then sharpens. It becomes both the shift in the forward process and the conditioning of the denoiser, so the reverse process starts near the noised prior and denoises a truncated chain (Figure~\ref{fig:overview}). The name reflects this composition, a graph diffusion feeding a denoising diffusion. Our contributions are as follows:

\begin{enumerate}
\item \textbf{A closed-form graph-heat prior as the backbone of the diffusion.} We derive a parameter-free preliminary forecast from the graph-heat equation~\cite{kondor_diffusion_2002}, and use it both to shift the forward process and to condition the denoiser, turning generation into refinement around the prior rather than synthesis from noise. This adapts residual-shift techniques from image restoration to spatio-temporal forecasting.

\item \textbf{A truncated reverse chain that cuts sampling cost.} Blending the graph-prior residual into the forward process following Resfusion~\cite{shi_resfusion_2024} lets the reverse chain begin from the noised prior partway along the schedule, so sampling runs only a truncated chain. Over the schedules we tested, measured sampling cost grows sublinearly with the scheduled length against roughly linearly for the evaluated full-chain baselines (Section~\ref{sec:latency}).
\item \textbf{DD-Net, a compact denoiser whose graph axis is gated automatically.} A single sub-million-parameter denoiser mixes along a spectral-temporal, a per-node channel, and a \emph{switchable} Chebyshev graph axis. An offline graph-spectral gate diagnostic reads the low-band energy of the prior residual and decides, per domain and from training data alone, whether that axis is needed: where the low-pass prior has already captured the spatial structure (air quality) the axis is switched off, removing its parameters and graph operations outright, and where high graph-frequency structure remains (traffic) it is switched on.
\end{enumerate}

The paper is structured as follows. Section~\ref{sec:system} develops the system, Section~\ref{sec:autoconf} derives the gate diagnostic, Section~\ref{sec:setup} gives the protocol, and Section~\ref{sec:results} reports accuracy, sampling cost, the ablation, and rolling revision in turn.

\section{Related Work}

\textbf{Spatio-temporal forecasting on sensor graphs.} Graph and attention models, STGCN~\cite{yu_spatio-temporal_2018}, DCRNN~\cite{li_diffusion_2018}, GMAN~\cite{zheng_gman_2020}, GMSDR~\cite{liu_msdr_2022}, ASTGCN~\cite{guo_attention_2019}, with automated search~\cite{wu_autocts_2022} and even linear models~\cite{zeng_are_2022}, are strong accuracy baselines but output point estimates. AirFormer~\cite{liang_airformer_2023} adds stochastic latent variables to capture forecast uncertainty, and a broad line of graph, recurrent and hybrid models targets pollutant concentration~\cite{wang_pm25-gnn_2021, tao_air_2019, qi_hybrid_2019, jin_multivariate_2021, zhang_hybrid_2021, xiao_dual-path_2022, zhou_theory-guided_2022, han_kill_2023}; see~\cite{zhang_systematic_2024-1} for a survey of the area. Being deterministic, the rest emit one trajectory and no predictive distribution.

\textbf{Stochastic spatio-temporal forecasting.} Denoising diffusion has become the most expressive probabilistic family on sensor graphs. TimeGrad~\cite{rasul_autoregressive_2021} applies denoising diffusion~\cite{ho_denoising_2020} to multivariate series and CSDI~\cite{tashiro_csdi_2021} to conditional imputation and forecasting, while DiffSTG~\cite{wen_diffstg_2023} and SpecSTG~\cite{lin_specstg_2024} target graph-structured forecasting; later work~\cite{hu_towards_2024} unifies spatio-temporal graph learning tasks under one diffusion model, and~\cite{LiuMingzhe2023PACD} applies the same machinery to spatio-temporal imputation. These models all run the full chain from a data-independent noise prior, so sampling cost is their shared weakness; accelerated samplers such as DDIM~\cite{song_denoising_2021} shorten a trained model's chain by striding its schedule, and TMDM~\cite{li_transformer_2024} pairs diffusion with a transformer. GDSS~\cite{jo_gdss_2022} instead generates graph \emph{structure} from a Gaussian prior. We take the graph as fixed, condition on the last observation, and start the reverse chain from a deterministic spectral prior partway along the schedule.

\textbf{Residual-shift diffusion and physics-guided forecasting.} Resfusion~\cite{shi_resfusion_2024}, ResShift~\cite{yue_resshift_2023} and RDDM~\cite{liu_residual_2024} build restoration diffusion paths around a degraded observation or its residual, Resfusion initializing the reverse process from the noised degraded input. Forecasting is given no such input, so we synthesize one in closed form. CorrDiff~\cite{mardani_corrdiff_2025} likewise refines a deterministic regression mean, but by a from-noise corrective diffusion; our residual instead enters the forward process itself. DYffusion~\cite{cachay_dyffusion_2023} and rolling weather models~\cite{erdm_2025} pursue related temporal dynamics, while physics-informed and neural-ODE models~\cite{raissi_physics_2017, hettige_airphynet_2024, ji_stden_2022, chen_neural_2019, rubanova_latent_2019, tian_air_2024} incorporate differential-equation constraints or learn continuous-time dynamics. We instead keep the equation fixed and let the generative stage absorb what it misses. The graph heat equation has a long history as a smoothing prior~\cite{kondor_diffusion_2002, shuman_emerging_2013}.

\textbf{Uncertainty.} MC Dropout~\cite{gal_dropout_2016} and DeepAR~\cite{salinas_deepar_2020} are standard probabilistic baselines, and uncertainty quantification in deep spatio-temporal models is active~\cite{wu_quantifying_2021}. Conformal prediction~\cite{stankeviciute_conformal_2021, vovk_algorithmic_2005} wraps any base forecaster in intervals with finite-sample marginal coverage under exchangeability, but presupposes such a predictor and adds a calibration step, whereas a diffusion model yields the predictive distribution natively. Our contribution relative to all of these is a systems view: one shared architecture whose spatial axis is set per domain by an offline diagnostic, fixing its operating point on the speed-accuracy-uncertainty trade-off.

\section{Problem Setting}

\begin{figure*}[t]
    \centering
    \includegraphics[width=0.58\textwidth]{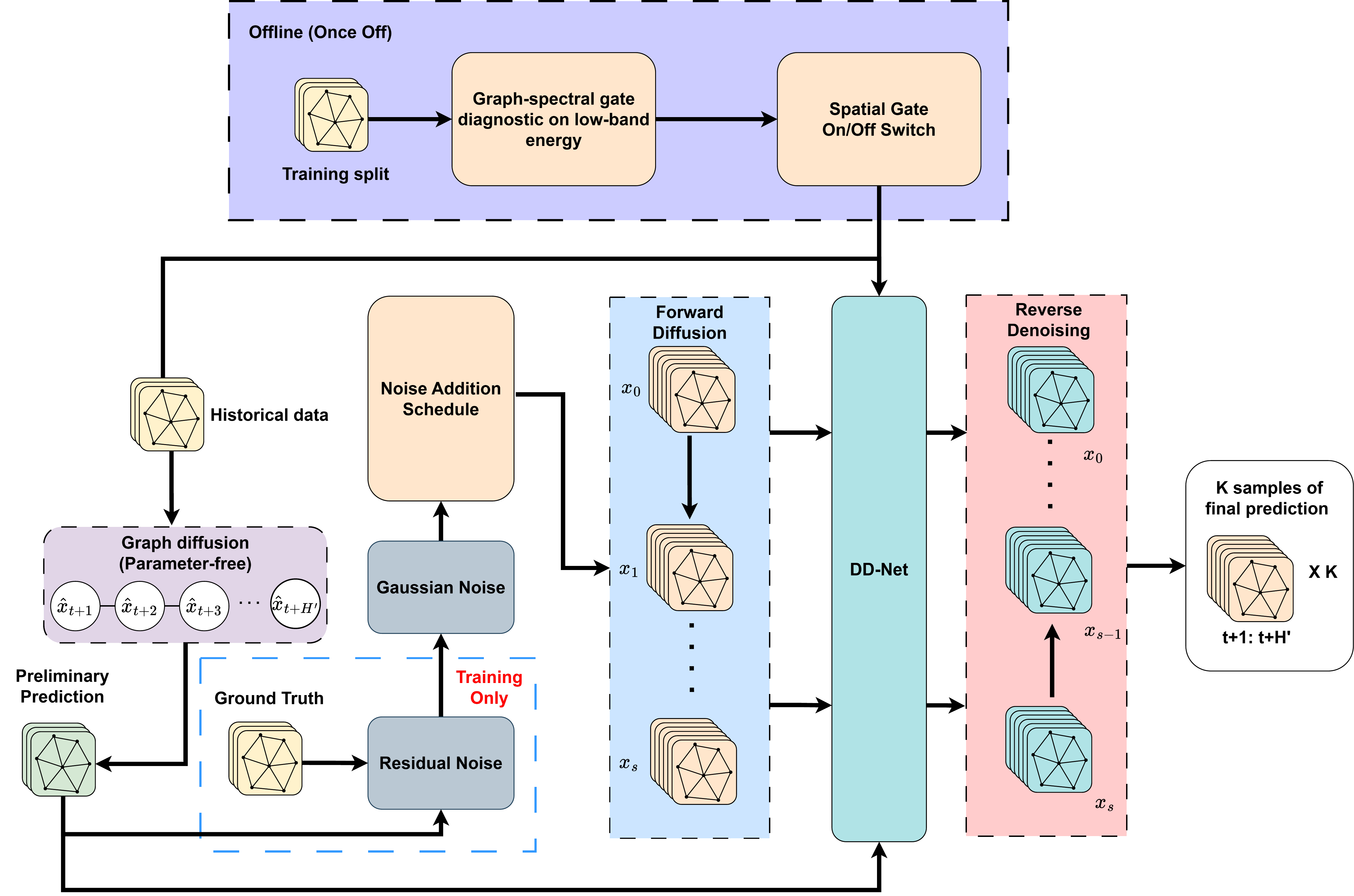}
    \Description{System diagram with an offline branch and a run-time branch. Offline, the graph-spectral gate diagnostic reads the low-band residual energy and switches the spatial gate on or off, which configures the denoiser. At run time, historical data passes through a parameter-free graph diffusion to a preliminary forecast; a forward diffusion chain corrupts the future while shifting it toward the prior residual, the shared DD-Net denoiser is applied, and a reverse denoising chain is sampled K times to produce the forecast distribution over the horizon. A residual-noise path is used during training only.}
    \caption{Double-Diffusion. The offline graph-spectral gate diagnostic (Section~\ref{sec:autoconf}) sets DD-Net's spatial gate; at run time the parameter-free graph-diffusion prior initializes a truncated denoising chain whose $K$ samples form the predictive distribution. The dashed residual-noise path is used only during training.}
    \label{fig:overview}
\end{figure*}

We consider $N$ sensors arranged on a graph $G$ whose adjacency comes from geographic proximity or the supplied road network, each carrying $C$ measured channels. Given $H$ past steps $x_{t-H+1:t} \in \mathbb{R}^{H \times N \times C}$, conventional spatio-temporal forecasting predicts the next $H'$ steps as the predictive distribution
\begin{equation}
p\!\left(x_{t+1:t+H'} \,\middle|\, x_{t-H+1:t}\right),
\label{eq:oneshot}
\end{equation}
reported through $K$ samples. The forecast is multivariate on every dataset: rather than a single target such as PM$_{2.5}$ or traffic flow, the model jointly predicts every measured channel (the pollutant species for air quality, and flow, occupancy, and speed for traffic).

In addition to this initial full-horizon forecast, real-time deployment requires the forecast to be refreshed as new readings arrive: once the first $m$ of the $H'$ horizon steps have been observed, the prediction of the remainder should be re-issued in the light of them. We therefore also consider a \emph{rolling} formulation that conditions on the revealed prefix,
\begin{equation}
p\!\left(x_{t+m+1:t+H'} \,\middle|\, x_{t-H+1:t},\, x_{t+1:t+m}\right),
\label{eq:rolling}
\end{equation}
which differs from Equation~\eqref{eq:oneshot} only by the revealed prefix. Both forecasts feed threshold and risk rules, so they must be full predictive distributions rather than single trajectories. Moreover, one agency typically runs several such networks across pollutants, road types, and cities, rewarding one architecture that transfers to a new domain from training data alone.

\section{The Double-Diffusion System}
\label{sec:system}

Figure~\ref{fig:overview} shows the full system. The name denotes a composition of two diffusions: a closed-form graph diffusion that produces a structural prior, and a learned denoising diffusion that refines it. The prior is not a separate model placed beside the denoiser; it is folded into the denoising diffusion in two roles: the residual between the prior and the future enters the forward process the model learns to invert, and the prior conditions the denoiser. As in any diffusion model, the denoiser still removes Gaussian noise, but it does so on a short chain that starts from the noised prior rather than from pure noise, following Resfusion~\cite{shi_resfusion_2024}.

\subsection{Graph diffusion prior}
On a sensor graph, nearby nodes co-vary because of physical coupling. We encode this inductive bias with the graph heat equation,
\begin{equation}
\frac{\partial X}{\partial t} = -kLX,
\label{eq:heat}
\end{equation}
where $L$ is the normalized graph Laplacian and $k > 0$ a diffusion coefficient (i.e., the smoothing rate). Its closed-form solution propagates the last observation $x_t$ forward over horizon step $h$:
\begin{equation}
\hat{x}(h) = U \, \mathrm{diag}\!\left(e^{-k \lambda_i h}\right) U^{\top} x_t,
\label{eq:prior}
\end{equation}
where $U$ and $\{\lambda_i\}$ are the eigenvectors and eigenvalues of $L$. Equation~\eqref{eq:prior} is a spectral low-pass forecast computed in two matrix products, with no learned parameters. It is deliberately not a simulator. It has no sources, sinks, or exogenous inputs. Its role is to shorten the distance the generative stage covers.

The coefficient $k$ sets how aggressively the prior smooths. As $k\to 0$ it reduces to persistence, the last observation held flat across the horizon; as $k$ grows it damps high graph-frequency modes faster and pulls the forecast toward the local graph average, and because the exponent scales with the horizon step $h$, distant steps are smoothed more than near ones. A moderate $k$ thus reproduces the spatial diffusion of a smooth field, such as a spreading pollutant plume, while leaving sharper structure for the generative stage to add; we fix it once as the shared development anchor, kept after the one-factor comparisons (Section~\ref{sec:ablation}), and reuse it unchanged across networks. The prior stays cheap because the eigendecomposition of $L$ is computed once per network and cached.

\subsection{Background: denoising diffusion}
We briefly review the DDPM formulation the system builds on. A DDPM~\cite{ho_denoising_2020} corrupts a clean sample $x_0$ over $S$ discrete steps by the forward Markov chain
\begin{equation}
q(x_s \mid x_{s-1}) = \mathcal{N}\!\left(x_s;\, \sqrt{1-\beta_s}\, x_{s-1},\, \beta_s \mathbf{I}\right),
\label{eq:ddpmfwd}
\end{equation}
with variance schedule $\{\beta_s\}$, which admits the closed form
\begin{equation}
q(x_s \mid x_0) = \mathcal{N}\!\left(x_s;\, \sqrt{\bar{\alpha}_s}\, x_0,\, (1-\bar{\alpha}_s) \mathbf{I}\right),
\quad \bar{\alpha}_s = \prod_{i=1}^{s} (1-\beta_i).
\label{eq:ddpmclosed}
\end{equation}
A neural denoiser $\epsilon_\theta$ is trained to predict the added noise with a mean squared error objective, and generation runs the learned reverse chain
\begin{equation}
p_\theta(x_{s-1} \mid x_s) = \mathcal{N}\!\left(x_{s-1};\, \mu_\theta(x_s, s),\, \sigma_s^2 \mathbf{I}\right)
\label{eq:ddpmrev}
\end{equation}
from pure noise $x_S$ down to $x_0$, one denoiser evaluation per step.

\subsection{Residual generative stage}
The generative stage is organized around the residual between the prior and the truth,
\begin{equation}
\mathit{Res} = \hat{x}_{t+1:t+H'} - x_{t+1:t+H'} \;\in\; \mathbb{R}^{H' \times N \times C},
\label{eq:residual}
\end{equation}
in the dataset-normalized signal space. The network does not generate this residual. Instead, following Resfusion~\cite{shi_resfusion_2024}, $\mathit{Res}$ enters as a drift in the forward process of Equation~\eqref{eq:ddpmfwd},
\begin{equation}
q(x_s \mid x_{s-1}, \mathit{Res}) = \mathcal{N}\!\left(x_s;\, \sqrt{1-\beta_s}\, x_{s-1} + \left(1 - \sqrt{1-\beta_s}\right) \mathit{Res},\, \beta_s \mathbf{I}\right),
\label{eq:blend}
\end{equation}
whose clean state is the future itself, so that marginally $x_s = \sqrt{\bar\alpha_s}\, x_{t+1:t+H'} + (1-\sqrt{\bar\alpha_s})\, \mathit{Res} + \sqrt{1-\bar\alpha_s}\, \epsilon$. The training target is neither the residual nor the noise alone but their combination,
\begin{equation}
\tilde{\epsilon} = \epsilon + \frac{\left(1-\sqrt{\alpha_s}\right)\sqrt{1-\bar\alpha_s}}{\beta_s}\, \mathit{Res},
\label{eq:epstilde}
\end{equation}
against which DD-Net is trained under a masked mean-squared error. During training each minibatch uniformly draws a revealed prefix $m \in \{0, 1, 3, 6, 12\}$: the first $m$ future steps are clamped to their observed values and excluded from the loss, so one trained model serves both evaluation views ($m{=}0$ initial, $m{=}6$ rolling). At inference the reverse chain starts from the noised prior and returns the forecast directly; the future appears in the clean state during supervised training only, as in any conditional diffusion model, and inference conditions only on the history and the prior (in the rolling view, also the revealed prefix). Because the drift shifts the forward marginal toward the prior as $s$ grows, Resfusion~\cite{shi_resfusion_2024} shows that the noised prior and the noised target approximately coincide in distribution at the intermediate step
\begin{equation}
S' = \arg\min_{s} \left| \sqrt{\bar{\alpha}_s} - \tfrac{1}{2} \right|,
\label{eq:sprime}
\end{equation}
so generation can begin there from the noised prior instead of from pure noise at step $S$. Resfusion computes $S'$ by this discrete minimization. However, for the linear schedule used here we derive, in Appendix~\ref{app:sprime}, the approximation $\sqrt{\bar\alpha_s} \approx \exp(-s^2\beta_S/4S)$ and with it the closed form
\begin{equation}
S' \approx \sqrt{4S\ln 2 \,/\, \beta_S},
\label{eq:sprimeclosed}
\end{equation}
which matches the exact discrete minimizer to under one step at every schedule we use; the small residual gap is the accumulated effect of the two approximations behind it (Appendix~\ref{app:sprime}). Two consequences follow. First, under the leading-order approximation $S'$ depends on $S$ and $\beta_S$ only through their ratio, and the equal-ratio schedules we test retain identical exact step counts (Section~\ref{sec:ablation}). Second, $S'$ grows approximately with the square root of the scheduled length, so lengthening the schedule raises DD's sampling cost sublinearly while the evaluated full-chain samplers pay for every step added; Section~\ref{sec:latency} confirms this growth rate by direct measurement. The acceleration is a literal truncation. The sampler runs only steps $S'$ down to $0$, and training draws $s$ from $\{1,\dots,S'\}$ to match. The start step $S'$ is fixed by the noise schedule and does not move with the prior; what a well-chosen coefficient $k$ buys is a smaller residual at this fixed start step, which lowers the variance of the denoiser's learning target. Resfusion was developed for image restoration, where the degraded input is given; bringing its residual-shift forward process to spatio-temporal graph forecasting, where that input is instead synthesized in closed form by the graph-heat prior, is to our knowledge new. This residual-shifted start places the method in the residual-diffusion family of ResShift~\cite{yue_resshift_2023} and RDDM~\cite{liu_residual_2024}.

\subsection{Denoiser architecture}
\label{sec:denoiser}
\begin{figure*}[t]
\centering
\includegraphics[width=0.56\textwidth]{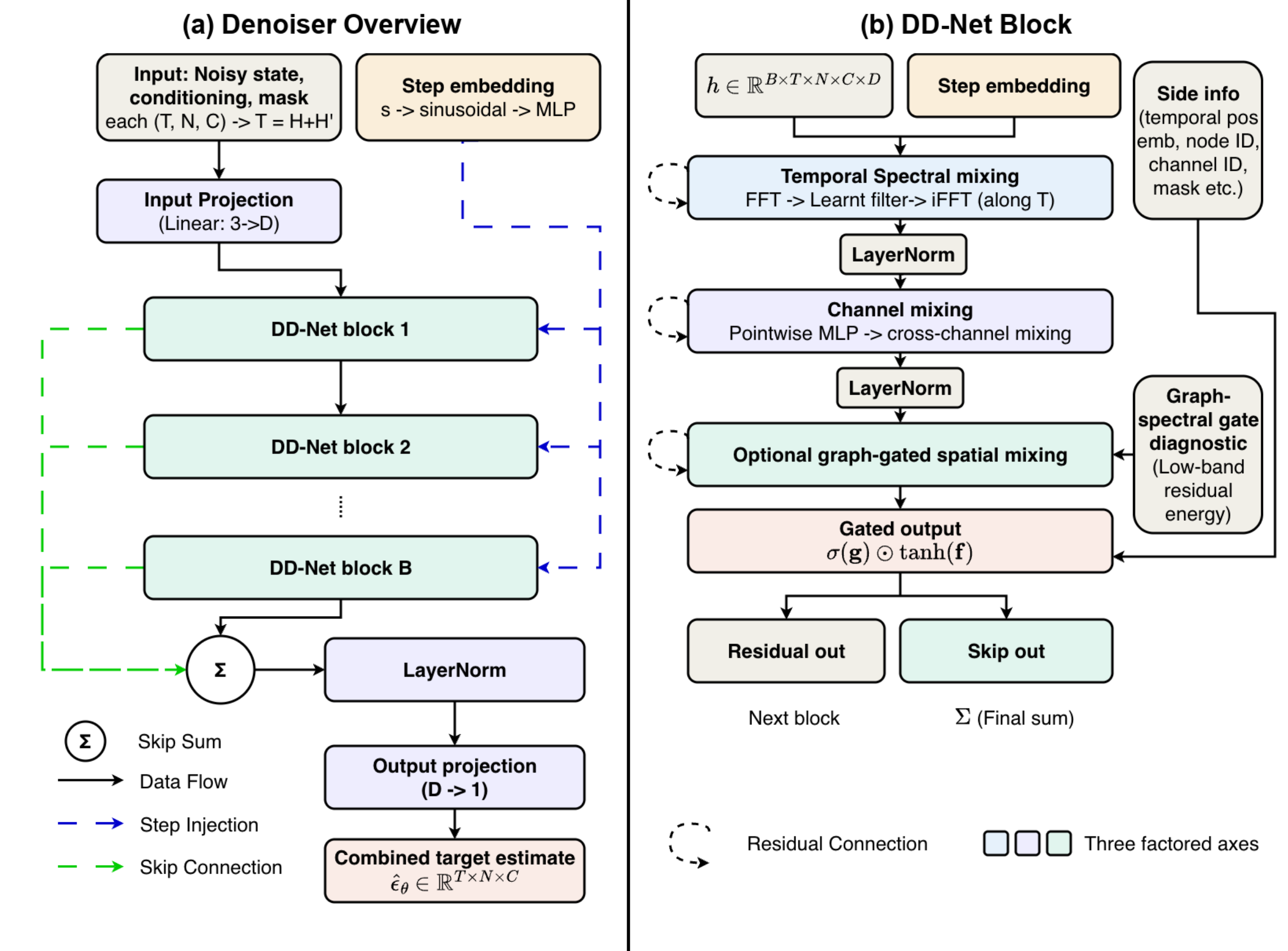}
\caption{DD-Net architecture. Each block combines spectral temporal mixing, per-node channel mixing, and optional Chebyshev graph-spatial mixing, with residual and skip connections. Section~\ref{sec:autoconf} sets the spatial gate per domain.}
\Description{A block diagram of the three-axis denoiser, showing spectral temporal mixing, per-node channel mixing, and an optional graph-gated spatial mixing stage with gated residual and skip connections.}
\label{fig:denoiser}
\end{figure*}

The denoiser, \emph{DD-Net} (Figure~\ref{fig:denoiser}), factorizes spatio-temporal denoising into three axes applied in sequence, each implemented as a gated residual block over a hidden width $D$ and stacked $B$ times.

\textbf{Input and conditioning.} The noisy state at diffusion step $s$, the conditioning signal (the history concatenated with the prior forecast), and the observation mask are stacked and linearly projected to $D$ channels, giving a hidden state of shape $(T, N, C, D)$ with $T = H + H'$. The diffusion step $s$ is encoded sinusoidally, passed through a small MLP, and added to the hidden state in every block. Positional, node, and channel identity embeddings and an observation mask are supplied to each block as side information (Figure~\ref{fig:denoiser}).

\textbf{Axis 1: spectral temporal mixing.} The hidden state is transformed along the time axis with a real-input fast Fourier transform and multiplied by learnable complex filters. A gate in the spirit of gMLP~\cite{liu_gmlp_2021} then modulates it (i.e., rescales each frequency): a small network reads the spectral magnitude and emits a multiplicative correction, so the spectrum is scaled by one plus that correction before the inverse transform. Every position therefore sees all $T$ steps at $O(T \log T)$ cost, in place of stacked convolutions or quadratic attention.

\textbf{Axis 2: channel mixing.} A per-node multilayer perceptron mixes the $C$ physical channels at each time step, capturing relationships such as the interaction between pollutant species or between flow, occupancy, and speed.

\textbf{Axis 3: spatial gate.} An optional Chebyshev graph convolution~\cite{defferrard_chebnet_2016} of order $r$ mixes each sensor with its $r$-hop neighborhood on $G$. Whether this gate is active is set per domain by the graph-spectral gate diagnostic of Section~\ref{sec:autoconf}; when the diagnostic switches it off, the Chebyshev branch is replaced by the identity map, so its parameters are never instantiated and no graph operation executes in any block.

Factorizing the three axes keeps the denoiser compact. Each mixes along a single mode of the $(T, N, C)$ tensor, so the parameter count grows additively rather than multiplicatively, and the network stays under a million parameters while reaching global temporal, full-channel and multi-hop spatial context. That suits the target: the residual the prior leaves is small and structured. Within each block, gated residual and skip connections let the network default to a near-identity map.

\section{The Graph-Spectral Gate Diagnostic}
\label{sec:autoconf}

The graph-heat prior is a low-pass filter on the sensor graph: it smooths the last observation across neighboring nodes and attenuates high graph-frequency structure, increasingly so at longer horizons~\cite{kondor_diffusion_2002, ntmaehara_lowpass_2019, wu_sgc_2019}. How much of the spatial signal it captures depends on the domain. Air-quality fields are smooth, since concentrations vary gradually between nearby monitors, so a graph diffusion reproduces most of the spatial pattern and leaves a small low-frequency residual. Traffic fields are abrupt: congestion onsets, bottlenecks and queue spillback (e.g., at a merge) make adjacent segments differ sharply, which is high graph-frequency structure a low-pass filter cannot represent, so the residual still carries spatial structure. DD-Net's spatial gate (Axis 3) can model that structure, but where the prior already captures the signal it adds parameters and inference-time graph operations for no gain. We therefore make the gate \emph{switchable} and set it per domain before any generative training, from one read-out of the prior residual.

The gate is a learnable Chebyshev graph convolution~\cite{defferrard_chebnet_2016}, able in principle to recover the high graph-frequency structure a fixed low-pass filter cannot~\cite{bo_beyond_2021, chien_gprgnn_2021}. Since it and the prior act on the same Laplacian spectrum, the only structure it can add is whatever the prior leaves in the residual. We therefore transform that residual, window by window over a domain's training split, into the graph spectral basis, $\tilde{r} = U^{\top} \mathit{Res}$, and measure the fraction of its energy in the lowest band,
\begin{equation}
\rho_G = \frac{\sum_{i \in \mathcal{B}_{\mathrm{low}}} \tilde{r}_i^{\,2}}{\sum_{i=1}^{N} \tilde{r}_i^{\,2}},
\qquad
\text{gate} =
\begin{cases}
\text{off}, & \rho_G > \tfrac{1}{2}, \\
\text{on}, & \text{otherwise},
\end{cases}
\label{eq:lowband}
\end{equation}
where $\mathcal{B}_{\mathrm{low}}$ is the lowest $\lfloor N/3 \rfloor$ modes by eigenvalue. The energies are summed rather than averaged: each training window's residual is projected onto the eigenbasis at the mid-horizon step, step $13$ of $24$, squared, and accumulated over windows and channels before the ratio is taken. The temporal control reported below is a different statistic, the energy fraction in the lower half of the one-sided rFFT bins across all horizon steps. When most of the residual energy already lies in this band we read the prior as having captured the spatial structure a graph filter could recover, and disable the gate. A large residual concentrated in one low mode would also score near one, so $\rho_G$ locates the residual's energy on the spectrum rather than measuring its magnitude. Where substantial energy remains at high graph frequencies, the gate is enabled instead. Like $S'$, the gate is fixed without forecast evaluation or per-domain search, here by one training statistic and one threshold.

\begin{figure}[t]
\centering
\includegraphics[width=0.8\linewidth]{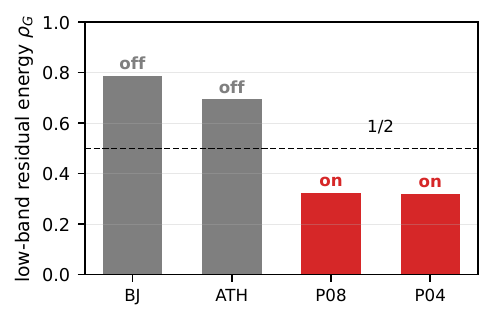}
\caption{The graph-spectral gate rule: the low-band residual energy fraction $\rho_G$ separates the domains at the one-half threshold (dashed), high for the air-quality networks (gate off) and low for the traffic networks (gate on).}
\Description{A bar chart of the low-band residual energy fraction for the four networks: high for the two air-quality networks and low for the two traffic networks, separated by the one-half threshold that sets the gate off or on.}
\label{fig:spectral}
\end{figure}

Applied to the four networks on the binary adjacency the models actually consume, the rule switches the gate off for both air quality domains ($\rho_G = 0.79$ and $0.69$) and on for both traffic domains ($0.32$ and $0.32$; Figure~\ref{fig:spectral}), and all four values sit at least $0.15$ from the threshold, so the decision is robust to moderate changes in the cutoff.

 The same statistic computed along the time axis is uniformly high across all four domains ($0.96$ to $0.98$) and does not separate them, so the discriminating structure is graph-spectral rather than a generic property of the residual. The smallest margin, $0.175$ on PEMS08, sits just above the $0.15$ bound, so the bound is close to tight.


\section{Experimental Setup}
\label{sec:setup}

\subsection{Datasets}
We evaluate on four real-world sensor networks, summarized in Table~\ref{tab:datasets}: two air quality networks (Beijing, Athens) and two traffic networks (PEMS08, PEMS04), whose graphs are built differently by family and then binarized.

For the air-quality networks we take a Gaussian kernel of pairwise geographic distance and zero weights below a fixed cutoff, so an edge is present when two monitors are close enough. For the traffic networks we take the road-network edge list supplied with the benchmark and apply no distance cutoff at all, since thresholding a sparse road graph would disconnect most of it. One topological property of PEMS04 is worth noting: the road-network edge list supplied with the benchmark is itself disconnected, splitting the graph into twelve components, so the graph-diffusion prior smooths within each component rather than across the whole network; the other three networks are connected. In both cases the models consume the binary support of the resulting adjacency. We treat the graph as a fixed input and do not vary this choice, and every graph-consuming model in Table~\ref{tab:main} is trained and scored on this same binary adjacency.

The air-quality networks predict pollutant concentrations, PM$_{2.5}$, PM$_{10}$, NO$_2$, and O$_3$ on both, with SO$_2$ and CO additionally on Beijing; the traffic networks forecast flow, occupancy, and speed. Each is forecast as a multivariate signal, with all $C$ channels predicted and scored jointly. The Athens network is a regional air-quality monitoring deployment~\cite{angelisRegionalDatasetsAir2024}; Beijing is the KDD Cup 2018 air-quality benchmark~\cite{kddcup_2018}, and PEMS08 and PEMS04 are the traffic benchmarks distributed with ASTGCN~\cite{guo_attention_2019}. While the air-quality networks are smaller and sampled hourly, the traffic networks are larger and sampled every five minutes, so the four span an order of magnitude in node count and two temporal resolutions (Table~\ref{tab:datasets}), making the per-domain gate decision a genuine transfer test.

We use $H{=}12$ input steps and $H'{=}24$ forecast steps throughout, with chronological 60/20/20 train, validation, and test splits and identical windows and normalization for every model. Preprocessing is uniform across methods. Gaps in the air-quality series are filled by a trailing mean of observed values and forward fill, with a leading-gap fallback fitted on the training split alone (zero for series never observed in training); imputed positions stay in training and evaluation for every model, so all methods are scored against the same field, and normalization statistics (the per-channel mean and scale) are fitted on training only.

\begin{table}[t]
\caption{The four sensor networks. All channels are jointly predicted and scored: six/four pollutants for Beijing/Athens and flow, occupancy, and speed for traffic.}
\label{tab:datasets}
\begin{tabular}{lllll}
\toprule
Domain & Dataset & $N$ & $C$ & Interval \\
\midrule
Air quality & Beijing & 32 & 6 & 1 hour \\
Air quality & Athens & 64 & 4 & 1 hour \\
Traffic & PEMS08 & 170 & 3 & 5 min \\
Traffic & PEMS04 & 307 & 3 & 5 min \\
\bottomrule
\end{tabular}
\end{table}

\subsection{Protocol and metrics}
Models train for up to 200 epochs with early stopping at patience 20, using Adam at learning rate $10^{-3}$ decayed by cosine to $10^{-5}$, no weight decay, and gradient clipping at $5$. Checkpoints are selected on the full validation split by sampled MAE at $K{=}1$; test evaluation is separate and uses $K{=}32$. Probabilistic models draw $K{=}32$ samples; the ensemble mean is scored for point error and the full ensemble for CRPS. For a forecast ensemble $\{\hat{y}^{(1)}, \dots, \hat{y}^{(K)}\}$ with mean $\bar{y}$ and a realized value $y$, averaged over all test positions:
\begin{equation}
\mathrm{MAE} = \mathbb{E}\left[ \left| \bar{y} - y \right| \right], \qquad
\mathrm{RMSE} = \sqrt{ \mathbb{E}\left[ \left( \bar{y} - y \right)^2 \right] },
\end{equation}
\begin{equation}
\mathrm{CRPS} = \mathbb{E}\left[ \frac{1}{K}\sum_{i=1}^{K} \left| \hat{y}^{(i)} - y \right| - \frac{1}{2K^2}\sum_{i=1}^{K}\sum_{j=1}^{K} \left| \hat{y}^{(i)} - \hat{y}^{(j)} \right| \right].
\end{equation}
All three metrics are computed per position (channel, node, and scored horizon step) and averaged over all positions, so one number per dataset summarizes all $C$ channels; the reported CRPS is thus a marginal score and does not evaluate cross-variable or cross-node dependence. MAE and RMSE are computed in raw measurement units (so for the traffic networks the reported figure is a mixed-unit average over flow, occupancy, and speed, dominated in scale by flow); CRPS~\cite{matheson_scoring_1976} is computed in normalized units. As the deterministic baselines emit no predictive distribution, we omit CRPS for them rather than score a point mass (i.e., a degenerate distribution). We fix $K{=}32$ samples for every probabilistic model, chosen as a realistic deployment budget rather than a convergence point; CRPS is reported at this fixed $K$, and the same $K$ is used in the latency measurements of Section~\ref{sec:results} so that accuracy and cost are reported at a single operating point.

\subsection{Baselines}
STGCN, DLinear, GMSDR, and GMAN (deterministic); DeepAR, MC Dropout~\cite{gal_dropout_2016, wu_quantifying_2021}, DiffSTG, and CSDI (probabilistic). All baselines run their reference implementations wrapped with a thin interface adapter for the multi-channel setting; GMSDR runs with two recurrent layers, and GMAN with one ST-attention block per encoder and decoder and a spectral spatial embedding. DiffSTG runs its published architecture at hidden size 64 with an $S{=}200$ chain and terminal noise $\beta_S{=}0.1$; CSDI~\cite{tashiro_csdi_2021} is adapted to forecasting on all four datasets at its default configuration, a 50-step from-noise chain with the quadratic noise schedule. We additionally report DiffSTG accelerated by DDIM sampling at 25 steps, which reuses the trained DiffSTG checkpoints unchanged and only swaps in the strided DDIM sampler, a uniform $25$-index stride over the $S{=}200$ schedule, at the stochasticity its reference implementation fixes, $\eta{=}1$ (no retraining; Appendix~\ref{app:ddim}).

\subsection{Rolling evaluation}
We report two views. The primary is the initial forecast, issued once and scored over the full horizon. The rolling view mirrors operational use: the first six steps of the horizon are revealed and only the remaining eighteen scored. Double-Diffusion implements this by resampling from the same history-anchored prior with the revealed prefix clamped at every reverse step, and scoring only the unrevealed suffix; the forecast is recomputed rather than patched, so the cost is that of a fresh draw. Every baseline is given the equivalent opportunity through a restart with a shifted input window. Within each view every model is scored on exactly the same target steps, and no model sees any step it is scored on. The rolling view instantiates the formulation of Equation~\eqref{eq:rolling} at $m{=}6$.

\subsection{Model configuration}
The same architecture and hyperparameter configuration is used across all four datasets, with a separate model trained per dataset: DD-Net at hidden width $D{=}64$ with $B{=}4$ blocks and spectral temporal mixing, under a million parameters. DD uses $S{=}100$ scheduled diffusion steps with terminal noise $\beta_S{=}0.1$, which fixes the Resfusion start step to $S'{=}52$: DD executes $52$ of its $100$ scheduled steps. The prior coefficient is fixed at $k{=}0.1$, the development anchor retained after the one-factor comparisons over $k\in\{0.05,0.1,0.2\}$ on the two probe domains (Section~\ref{sec:ablation}), and applied unchanged to all four datasets. DD's spatial gate, when active, uses Chebyshev order $r{=}3$, with its per-domain on/off setting fixed by the diagnostic of Section~\ref{sec:autoconf}. Section~\ref{sec:ablation} examines each of these choices one factor at a time.

\section{Results}
\label{sec:results}

\begin{table*}[t]
\caption{Full-test results at $K{=}32$ (three-seed means). \emph{Initial} scores steps 1--24; \emph{Rolling} reveals six steps and scores steps 7--24 with every method given the revealed prefix. \textbf{Bold} and \underline{underlining} mark the best and second-best result within each dataset, metric, and view.}
\label{tab:main}
\footnotesize\setlength{\tabcolsep}{3.2pt}
\begin{tabular}{l rrr rrr rrr rrr}
\toprule
 & \multicolumn{3}{c}{Beijing} & \multicolumn{3}{c}{Athens} & \multicolumn{3}{c}{PEMS08} & \multicolumn{3}{c}{PEMS04} \\
\cmidrule(lr){2-4}\cmidrule(lr){5-7}\cmidrule(lr){8-10}\cmidrule(lr){11-13}
Model & MAE & RMSE & CRPS & MAE & RMSE & CRPS & MAE & RMSE & CRPS & MAE & RMSE & CRPS \\
\midrule
\multicolumn{13}{l}{\emph{Initial (steps 1 to 24)}} \\
STGCN & 21.83 & 53.85 & n/a & 10.05 & 16.94 & n/a & 8.81 & 21.06 & n/a & 10.74 & 26.18 & n/a \\
DLinear & 21.32 & 53.49 & n/a & 9.45 & 16.57 & n/a & 11.14 & 27.27 & n/a & 13.63 & 33.02 & n/a \\
GMSDR & 22.74 & 53.34 & n/a & 9.97 & 17.09 & n/a & 7.26 & 17.66 & n/a & 8.56 & 21.15 & n/a \\
GMAN & 21.87 & 53.55 & n/a & 9.68 & 17.72 & n/a & 7.06 & 17.25 & n/a & 8.52 & 21.11 & n/a \\
DeepAR & 22.63 & 54.67 & 0.378 & 10.52 & 18.43 & 0.397 & 9.18 & 23.49 & 0.192 & 9.86 & 25.47 & 0.180 \\
MC Dropout & 22.23 & \textbf{52.35} & 0.418 & 9.52 & 16.52 & 0.405 & 7.56 & 18.06 & 0.161 & 9.62 & 23.27 & 0.168 \\
CSDI & \underline{20.98} & \underline{52.51} & 0.355 & \textbf{8.65} & \textbf{15.97} & \underline{0.355} & \textbf{5.80} & \textbf{14.91} & \textbf{0.125} & \textbf{7.69} & \textbf{19.50} & \underline{0.138} \\
DiffSTG & 21.68 & 53.01 & \underline{0.348} & 11.06 & 19.06 & 0.416 & 8.75 & 20.91 & 0.165 & 15.04 & 33.93 & 0.214 \\
DiffSTG (DDIM25) & 21.83 & 53.36 & 0.354 & 11.08 & 19.13 & 0.421 & 9.79 & 22.86 & 0.181 & 17.57 & 38.40 & 0.246 \\
\textbf{DD} & \textbf{20.68} & 54.29 & \textbf{0.321} & \underline{8.90} & \underline{16.12} & \textbf{0.327} & \underline{6.16} & \underline{15.63} & \underline{0.128} & \underline{7.97} & \underline{20.42} & \textbf{0.134} \\
\midrule
\multicolumn{13}{l}{\emph{Rolling (6 steps revealed, scored on steps 7 to 24)}} \\
STGCN & 20.63 & 51.86 & n/a & 9.83 & 16.72 & n/a & 8.24 & 19.71 & n/a & 9.95 & 24.40 & n/a \\
DLinear & 20.04 & 51.52 & n/a & 9.31 & 16.51 & n/a & 9.61 & 23.76 & n/a & 11.80 & 28.91 & n/a \\
GMSDR & 21.87 & 51.82 & n/a & 9.55 & 16.53 & n/a & 6.98 & 16.97 & n/a & 8.23 & 20.36 & n/a \\
GMAN & 20.30 & 51.14 & n/a & 9.31 & 17.53 & n/a & 6.77 & 16.56 & n/a & 8.19 & 20.35 & n/a \\
DeepAR & 21.54 & 53.13 & 0.361 & 10.09 & 17.91 & 0.381 & 8.85 & 22.64 & 0.187 & 9.55 & 24.76 & 0.175 \\
MC Dropout & 21.08 & \textbf{50.18} & 0.398 & 9.14 & 16.08 & 0.388 & 7.21 & 17.27 & 0.153 & 9.03 & 21.97 & 0.158 \\
CSDI & \underline{19.76} & \underline{50.87} & 0.335 & \underline{8.30} & \underline{15.59} & \underline{0.340} & \textbf{5.58} & \textbf{14.33} & \textbf{0.118} & \textbf{7.40} & \textbf{18.81} & \underline{0.131} \\
DiffSTG & 20.59 & 51.25 & \underline{0.331} & 10.76 & 18.79 & 0.405 & 7.79 & 18.81 & 0.151 & 13.21 & 29.64 & 0.191 \\
DiffSTG (DDIM25) & 20.81 & 51.72 & 0.338 & 10.82 & 18.88 & 0.411 & 8.67 & 20.38 & 0.165 & 15.98 & 34.81 & 0.224 \\
\textbf{DD} & \textbf{19.66} & 54.11 & \textbf{0.303} & \textbf{8.22} & \textbf{15.24} & \textbf{0.302} & \underline{5.78} & \underline{14.83} & \underline{0.121} & \underline{7.42} & \underline{19.21} & \textbf{0.124} \\
\bottomrule
\end{tabular}
\end{table*}

\subsection{Accuracy and probabilistic scores}
Table~\ref{tab:main} presents the main comparison. First, on CRPS (the proper probabilistic score that jointly rewards sharp and accurate distributions), DD attains the lowest value on Beijing, Athens and PEMS04 and CSDI on PEMS08, with DeepAR and MC Dropout behind both throughout; the ranking is identical in the initial and rolling views. Second, on point error no family dominates. The deterministic models are strong point predictors but emit a single trajectory. However, DD and CSDI, the two strongest probabilistic methods, stay competitive with them on MAE while supplying the full predictive distribution.

On the two air-quality networks the probabilistic pair is at its strongest. DD and CSDI improve on every deterministic baseline's MAE, and DD holds the lowest CRPS on both networks, by $9.5\%$ on Beijing and $11.2\%$ on Athens over CSDI in the rolling view. RMSE is the exception. DD trails the deterministic block on Beijing RMSE in both views, and the best Beijing RMSE overall belongs to MC Dropout.

On the traffic networks the deterministic baselines close much of that gap, GMAN and GMSDR in particular. Nevertheless, DD stays ahead of them on point error (PEMS08 initial $6.16$ MAE against GMAN's $7.06$ and GMSDR's $7.26$; PEMS04 $7.97$ against $8.52$ and $8.56$) while adding the distribution they lack. Among the other probabilistic baselines, MC Dropout trails DD by $1.40$ and $1.65$ MAE on the two traffic networks and DeepAR by $3.02$ and $1.89$.

CSDI is therefore the baseline to beat. The closest contest is PEMS08. There CSDI leads DD's CRPS by $2.8\%$ in the initial view and $1.9\%$ in the rolling (on the unrounded means of Table~\ref{tab:seeds}); the Welch test leaves the gap unresolved ($p{=}0.113$, Table~\ref{tab:sig}), and we report it as a consistent CSDI lead. Where DD leads instead, the margins are wider than where it trails. On point error the two trade the lead and no gap exceeds $0.36$ MAE: DD is lower on Beijing ($20.68$ against $20.98$ in the initial view) and Athens rolling, CSDI on PEMS08 ($5.80$ against $6.16$), PEMS04 ($7.69$ against $7.97$) and Athens initial. The overall pattern is the one the systems view anticipates: a strong closed-form prior followed by a truncated residual diffusion matches or beats the best point predictors on MAE while retaining a full predictive distribution.

The DDIM25 rows follow their parent checkpoints at matched seeds, nearly exactly on the air-quality networks and $7$ to $20\%$ worse on traffic (per-seed pairing in Appendix~\ref{app:ddim}).

\subsection{Sampling cost and the speed-accuracy trade-off}
\label{sec:latency}
Sampling cost is where DD's truncated chain pays. It manifests in two aspects: a constant factor at each method's own operating point, and a growth rate. We take the growth rate first.

\begin{figure}[t]
\centering
\includegraphics[width=\linewidth]{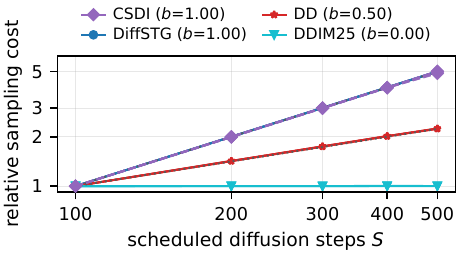}
\caption{Sampling-cost growth against scheduled chain length $S$ on logarithmic axes, all four networks overlaid per model, each series normalized at its own $S{=}100$ cost. The legend reports the fitted $b$ in $T \propto S^{b}$; guides show linear and square-root growth, and the CSDI and DiffSTG series overlap. These are matched-schedule cost curves rather than the operating points of Table~\ref{tab:latency}.}
\Description{A log-log plot of normalized sampling cost against scheduled diffusion steps for four models over four networks; sixteen curves collapse onto three lines, one following a square-root reference slope, two following a linear slope, and one flat.}
\label{fig:scaling}
\end{figure}

\textbf{Over the schedules we tested, the truncation changes the growth rate of sampling cost itself.} A scheduled chain length $S$ fixes the corruption process. CSDI and DiffSTG reverse the whole chain and execute all $S$ steps, DDIM25 strides the schedule and always executes $25$, and DD executes only the $S'$ steps derived from the scheduled $S$ (Equation~\ref{eq:sprimeclosed}). To measure the consequence, we time all four samplers on a common scheduled $S \in \{100,\dots,500\}$ at $K{=}32$, batch size $16$, one discarded warmup and the median of three repeats, under one fp32 precision policy, and fit $T \propto S^{b}$, where $T$ is the sampling time per window and $b$ the fitted growth exponent (Figure~\ref{fig:scaling}, Table~\ref{tab:scaling}). On the four networks DD returns $b$ between $0.501$ and $0.505$, against $0.990$ to $1.002$ for CSDI and $0.999$ to $1.002$ for DiffSTG. In Figure~\ref{fig:scaling} the DD series lie on the square-root guide while the CSDI and DiffSTG series lie on the linear guide and nearly coincide, which is why only three bands are visible. The exponent is a measured growth rate over the tested range rather than an asymptotic complexity class: the closed form approximates the implemented schedule, and raising $S$ at fixed $\beta_S$ changes the corruption process rather than refining one trajectory. As a result the gap compounds with the schedule rather than staying put: the matched-schedule CSDI-to-DD ratio rises from $3.83\times$ at $S{=}100$ to $8.63\times$ at $S{=}500$ on Beijing ($16.49$ against $1.91$\,s per window), with endpoints of $7.89\times$ on Athens, $5.01\times$ on PEMS08 and $4.69\times$ on PEMS04. The ordering follows graph size: on a larger network DD-Net's per-step cost approaches CSDI's, leaving the step-count ratio as the margin. The prior therefore buys speed in one precise sense: while no single denoiser evaluation becomes cheaper, the number of required evaluations grows with the square root of the schedule instead of linearly with it over this range. The sweep, however, holds each trained model fixed and measures sampling speed only. Accuracy at longer schedules is untested.

Table~\ref{tab:latency} turns to the second aspect, the cost each method pays at its own operating point, timed fresh at $K{=}32$ on one H100 NVL in fp32 with matmul TF32 disabled and cuDNN TF32 enabled; the closed-form prior itself costs about $0.01$\,ms.

\begin{table}[t]
\caption{Amortized sampling time per window (ms), $K{=}32$, one H100 NVL (effective batch $64$; median of three post-warmup repeats). Values measure batched throughput rather than per-request latency. DD executes $52$ of $100$ scheduled steps; DDIM25 timing uses $\eta{=}0$.}
\label{tab:latency}
\small\setlength{\tabcolsep}{4.5pt}
\begin{tabular}{l r rrrr}
\toprule
Model & Steps & Beijing & Athens & PEMS08 & PEMS04 \\
\midrule
DiffSTG (DDIM25) & 25 & 216 & 429 & 1387 & 2683 \\
DD & 52\,/\,100 & 870 & 1168 & 3476 & 6760 \\
CSDI & 50 & 1516 & 2000 & 4000 & 7144 \\
DiffSTG & 200 & 1727 & 3429 & 11099 & 21826 \\
\bottomrule
\end{tabular}
\end{table}

At the operating points, DD is faster than both full-chain samplers on every network: $1.06\times$ to $1.74\times$ against CSDI, a geometric mean of $1.38\times$, and from $1.98\times$ on Beijing to $3.23\times$ on PEMS04 against DiffSTG. Two effects set the constant: (1) DD executes $52$ steps to CSDI's $50$ and DiffSTG's $200$, and (2) the inclusive amortized cost per executed step (i.e., including ensemble chunking and sampling overhead) is lower for DD-Net than for CSDI's attention denoiser and higher than for the DiffSTG U-Net. Against CSDI the near-equal step counts leave the per-step gap as the whole margin, which fades with graph size, from $1.74\times$ on Beijing to $1.06\times$ on PEMS04; against DiffSTG the four-to-one step ratio dominates everywhere. None of this touches the exponent. The operating point moves the constant, while the truncation sets the slope.

\begin{figure*}[t]
\centering
\includegraphics[width=0.84\textwidth]{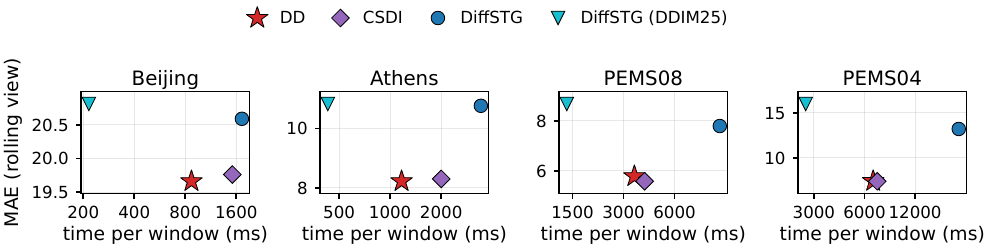}
\caption{Operating-point speed--accuracy trade-off: rolling MAE (Table~\ref{tab:main}) against amortized sampling time (Table~\ref{tab:latency}, log axis); lower left is better. DDIM25 combines $\eta{=}1$ accuracy with $\eta{=}0$ timing; both execute 25 steps.}
\Description{Four scatter panels, one per network, each plotting four models by sampling time on a log axis against rolling MAE. DD appears lowest and left on the frontier in every panel.}
\label{fig:pareto}
\end{figure*}

Figure~\ref{fig:pareto} places accuracy against this measured operating-point cost: DD sits on the speed-accuracy frontier of every panel, the only cheaper sampler is well above it in error, and the methods that approach its accuracy pay an equal or larger sampling bill.

\subsection{Ablation and hyperparameter analysis}
\label{sec:ablation}
We ablate on PEMS08 and Beijing, the two domains the configuration was developed on; Athens and PEMS04 received the frozen configuration without retuning, so their rows in Table~\ref{tab:main} are held-out transfers. Table~\ref{tab:ablation} reports the study at full scale: every cell is trained under the same protocol as Table~\ref{tab:main} and scored on the complete test split at $K{=}32$, one factor at a time from a shared anchor, the development configuration $(S,\beta_S,k,D,B,r)=(100,\,0.2,\,0.1,\,64,\,4,\,3)$, where $D$ is DD-Net's hidden width (the model dimension of Section~\ref{sec:denoiser}), $B$ its block count, and $r$ the Chebyshev order. Training used the train split only; selection among cells used full-test scores under our pre-specified rule, adopting a setting when it wins at least two of MAE, RMSE and CRPS (RMSE is omitted from Table~\ref{tab:ablation} for space), so the two probe domains are test-informed. The spatial-gate setting is not taken from this table at all, being fixed by the offline diagnostic of Section~\ref{sec:autoconf} before any training.

\begin{table}[t]
\caption{One-factor study on Beijing and PEMS08 (full test, $K{=}32$, initial view; one seed per cell). Rows vary from the anchor $(S,\beta_S,k,D,B,r)=(100,\,0.2,\,0.1,\,64,\,4,\,3)$ and are comparable only within groups. \textbf{Bold} marks the shipped setting; $S'$ is the executed start step. The diagnostic of Section~\ref{sec:autoconf} fixes the gate off for Beijing and on for PEMS08.}
\label{tab:ablation}
\footnotesize\setlength{\tabcolsep}{4pt}
\begin{tabular}{llrrrrr}
\toprule
 & & & \multicolumn{2}{c}{Beijing} & \multicolumn{2}{c}{PEMS08} \\
\cmidrule(lr){4-5}\cmidrule(lr){6-7}
Group & Setting & $S'$ & MAE & CRPS & MAE & CRPS \\
\midrule
Spatial gate & off (sp0) &  & \textbf{20.72} & \textbf{0.323} & 7.30 & 0.143 \\
 & on (sp1) &  & 20.41 & 0.325 & \textbf{6.11} & \textbf{0.130} \\
\midrule
Width $D$ & 16 &  & 21.05 & 0.329 & 8.20 & 0.164 \\
 & 32 &  & 21.02 & 0.332 & 7.40 & 0.149 \\
 & \textbf{64} &  & \textbf{20.72} & \textbf{0.323} & \textbf{6.11} & \textbf{0.130} \\
\midrule
Depth $B$ & 2 &  & 20.55 & 0.329 & 6.94 & 0.141 \\
 & \textbf{4} &  & \textbf{20.72} & \textbf{0.323} & \textbf{6.11} & \textbf{0.130} \\
 & 6 &  & 20.63 & 0.321 & 6.03 & 0.129 \\
\midrule
Graph order $r$ & 2 &  & -- & -- & 6.15 & 0.128 \\
 & \textbf{3} &  & -- & -- & \textbf{6.11} & \textbf{0.130} \\
 & 4 &  & -- & -- & 6.21 & 0.131 \\
\midrule
Prior coeff.\ $k$ & 0.05 &  & 20.38 & 0.321 & 6.15 & 0.129 \\
 & \textbf{0.1} &  & \textbf{20.72} & \textbf{0.323} & \textbf{6.11} & \textbf{0.130} \\
 & 0.2 &  & 20.61 & 0.327 & 6.31 & 0.130 \\
\midrule
Schedule $(S,\beta_S)$ & $\phantom{0}50,\,0.2$ & 26 & 20.84 & 0.330 & 6.30 & 0.132 \\
 & $100,\,0.4$ & 26 & 20.90 & 0.332 & 6.30 & 0.132 \\
 & $100,\,0.2$ & 37 & 20.72 & 0.323 & 6.11 & 0.130 \\
 & $\mathbf{100,\,0.1}$ & \textbf{52} & \textbf{20.10} & \textbf{0.313} & \textbf{6.04} & \textbf{0.127} \\
 & $200,\,0.2$ & 52 & 20.12 & 0.313 & 6.07 & 0.126 \\
\bottomrule
\end{tabular}
\end{table}

Every cell is a single seed, and the rows outside the schedule group sit at the $\beta_S{=}0.2$ anchor; we therefore read the table against seed noise rather than as a ranking. Three findings stand out.

\textbf{Two components dominate.} Width is the largest effect: $D{=}64$ improves PEMS08 MAE by $1.29$ over $D{=}32$ and $2.09$ over $D{=}16$, seven and twelve times the shipped model's three-seed standard deviation, and it wins all three metrics on both domains. The spatial gate is second, worth $1.19$ MAE on PEMS08 ($6.11$ against $7.30$), again on all three. Graph order is not in this class, as $r \in \{2,3,4\}$ spans $0.10$ MAE on PEMS08, and the prior coefficient behaves like graph order: $k{=}0.1$ is the better setting on PEMS08 while $k{=}0.05$ is marginally lower on Beijing, a disagreement of $0.34$ MAE that sits under half the seed standard deviation, so one value is carried to every domain rather than tuning $k$ per network. Depth splits, with two blocks worse than four by $0.83$ MAE on PEMS08 and six blocks close enough to be settled only by the head-to-head below.

\textbf{The ablation corroborates the gate diagnostic.} We set the gate by the residual-energy rule of Section~\ref{sec:autoconf} (Figure~\ref{fig:spectral}), which switches it off on air quality because the prior residual's spatial energy is already concentrated in the low band. The table bears this out empirically. The traffic half is unambiguous, the gate winning all three metrics on PEMS08, while on Beijing the effect splits by metric (turning it on improves MAE by $0.31$ but worsens RMSE by $1.19$ and CRPS by $0.002$), consistent with the off decision for a domain whose spatial structure the prior already carries.

\textbf{The schedule reduces to one number.} Under the closed form (Equation~\ref{eq:sprimeclosed}) $S'$ depends on $S$ and $\beta_S$ only through their ratio, and both equal-ratio pairs in the sweep bear this out: $(100,\,0.1)$ and $(200,\,0.2)$ both retain $52$ steps, and $(50,\,0.2)$ and $(100,\,0.4)$ both retain $26$. These are separately trained models with different corruption processes rather than one process reparameterized; what we observe is that matched retained-step counts give closely matched scores. The $S'{=}52$ pair differs by $0.02$ MAE on Beijing and $0.03$ on PEMS08, the $S'{=}26$ pair by $0.06$ and less than $0.01$, against a group spread of $0.80$ and $0.26$. Matched-$S'$ configurations consequently agree an order of magnitude more tightly than the group varies, which is why we report one schedule parameter rather than two. We ship $(100,\,0.1)$, the schedule we trained and confirmed; both members of the pair execute $52$ evaluations, so the pair is matched in sampling cost. Across the $S' \in [26,52]$ its schedule group covers, the Beijing spread is $0.80$ MAE against a seed standard deviation of $0.75$, so the response is close to flat above $S'{\approx}30$ and $S'{=}52$ is a confirmed setting rather than a measured optimum.

A row here is a one-factor change from the anchor, so a within-group winner has beaten the \emph{anchor} rather than the shipped cell; compared directly under the pre-specified rule, the shipped configuration beats every challenger on both probe domains. The two closest contests, $(200,\,0.2)$ and six blocks, lose only single metrics by margins far inside the seed noise; a CRPS-first rule would hand PEMS08 to $(200,\,0.2)$, the one pair on which the two rules diverge.

\subsection{When rolling revision helps}
\label{sec:rolling-when}
\begin{figure}[t]
\centering
\includegraphics[width=0.72\linewidth]{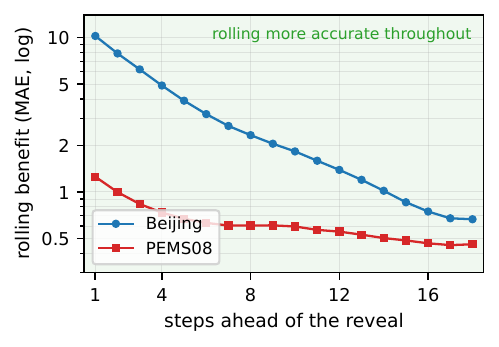}
\caption{Per-step rolling benefit (initial-view MAE minus rolling-view MAE at matched steps 7--24) against distance past the six-step reveal, on a log ordinate ($K{=}32$, seed 0, first 640 test windows). Positive values favor rolling and mainly reflect the shorter effective lead; subset error levels are not comparable with the full-split Table~\ref{tab:main}.}
\Description{A line plot of the per-step rolling benefit for Beijing and PEMS08 against steps ahead of the reveal, large near the reveal and decaying with distance while remaining positive at every scored step.}
\label{fig:rolling}
\end{figure}

Table~\ref{tab:main} reports the rolling view under a single six-step reveal. Figure~\ref{fig:rolling} resolves that gain step by step. Both arms forecast the same twenty-four-step window and are scored on the same horizon steps 7 to 24; the per-step benefit is their MAE difference at a matched target step, in raw units. The benefit is largest immediately past the reveal, at $10.24$ MAE on the first Beijing step and $1.25$ on PEMS08, declines overall with distance, and remains positive at every scored step, its smallest values being $0.66$ at eighteen steps out on Beijing and $0.45$ at seventeen on PEMS08.

Almost all of this gain is a shorter effective lead time rather than a revision-specific mechanism. Because the revealed prefix is clamped to the observations, the rolling arm reaches any target having seen readings six steps closer to it. Therefore its effective lead (i.e., its gap to the target) is simply shorter. Differencing the initial-view curve against itself at matched lead, its MAE at step $6{+}d$ minus its MAE at step $d$, reproduces $93.9\%$ of the mean benefit on Beijing and $84.1\%$ on PEMS08; the residual at matched effective lead is small and positive on both, $+0.18$ and $+0.10$ MAE. As the prior is built from the history window alone and is never re-anchored on the revealed prefix, the rolling arm carries a staler prior at matched lead, so the small positive residual persists despite that handicap. Establishing a refinement mechanism beyond fresher conditioning would need a control holding conditioning fixed while varying re-denoising depth; on the evidence here, the gain is fresher conditioning. For deployment the reading is simple: re-issuing the forecast as readings arrive helps, the help concentrates on the imminent steps, and at no scored step did it hurt.

\section{Limitations and future work}
The gate diagnostic turns an observation into an inference-time saving: where the low-pass prior has already captured the spatial structure, as on the air-quality networks, the spatial axis is switched off outright, with its parameters and graph operations removed (Section~\ref{sec:autoconf}). The decision currently rests on one residual-energy fraction against one threshold, and extending it is a natural next step; in particular, a within-domain study varying graph size at fixed domain would characterize the rule beyond a suite in which the smaller graphs ($N{=}32$ and $64$) carry air quality and the larger ($N{=}170$ and $307$) carry traffic; we leave this extension to future work.

Although the truncated chain lowers how sampling cost grows with the schedule, it does not make DD cheap in absolute terms: the reverse process still runs $52$ of $100$ scheduled steps rather than a single-digit count. DD is therefore faster than the full-chain diffusion forecasters but not than the strided 25-step sampler of Appendix~\ref{app:ddim} or the deterministic baselines, which run in one forward pass. Pushing below the Resfusion start step, for example with few-step probability-flow solvers, is left to future work.

We summarize probabilistic quality with CRPS but certify no explicit interval coverage; a conformal pass~\cite{stankeviciute_conformal_2021, vovk_algorithmic_2005} would add it under exchangeability assumptions. Finally, the gate is a binary decision at a fixed threshold: a continuous gain, a per-node filter, or a gate conditioned on the noise level could adapt more finely, but each trades away the parameter-free transferability that makes the present rule attractive across a fleet of heterogeneous networks.

Taken together the results suggest a recipe for a new network: compute the prior and its residual spectrum, set the gate from the one-half rule, and train the shared configuration, with no per-site search. At run time, serve the initial forecast for planning and overlay rolling updates for the imminent steps, from the same model. A $K{=}32$ ensemble costs roughly $0.9$ to $6.8$ seconds per window on one accelerator (Table~\ref{tab:latency}), ample headroom against sampling intervals of five minutes to an hour.

\section{Conclusion}

Double-Diffusion integrates a closed-form graph-heat prior into a denoising diffusion model in two roles, shifting the forward process and conditioning the denoiser, so that generation refines the prior over a truncated chain; an offline graph-spectral gate diagnostic then sets DD-Net's spatial axis per domain from training data alone. Across four urban sensor networks Double-Diffusion attains the lowest observed mean CRPS among the evaluated probabilistic methods on three of the four, ranks second within 3\% of the best on the fourth, and stays competitive in point accuracy. Moreover, truncating the reverse chain at the Resfusion start step, together with DD-Net's low per-step cost, makes DD faster than the evaluated full-chain diffusion forecasters at their own operating points. The more durable statement concerns scaling: over schedules $S{=}100$ to $500$, DD's sampling cost grows with the square root of the scheduled length while the full-chain baselines grow linearly, so the advantage widens as schedules lengthen, reaching $8.6\times$ over CSDI at $S{=}500$ on the smallest network in the cost-only benchmark.

\phantomsection\label{pg:acks}
\begin{acks}
We thank ResetData for providing the H200 GPUs used in this work. We used large language models for code refactoring, figure generation, and manuscript editing.
\end{acks}

\clearpage
\phantomsection\label{pg:refs}
\bibliographystyle{ACM-Reference-Format}
\bibliography{references}

@misc{kddcup_2018,
  title = {{KDD Cup} 2018},
  author = {{ACM SIGKDD}},
  year = {2018},
  url = {https://www.kdd.org/kdd2018/kdd-cup},
  lastaccessed = {August 31, 2026},
}

@inproceedings{angelisRegionalDatasetsAir2024,
	address = {Athens, Greece},
	title = {Regional Datasets for Air Quality Monitoring in European Cities},
	doi = {10.1109/IGARSS53475.2024.10640879},
	booktitle = {{IGARSS} 2024 - 2024 {IEEE} {International} {Geoscience} and {Remote} {Sensing} {Symposium}},
	publisher = {IEEE},
	author = {Angelis, G.-F. and Emvoliadis, A. and Theodorou, T.-I. and Zamichos, A. and Drosou, A. and Tzovaras, D.},
	year = {2024},
	pages = {6875--6880},
}

@article{han_kill_2023,
	title = {Kill Two Birds With One Stone: A Multi-View Multi-Adversarial Learning Approach for Joint Air Quality and Weather Prediction},
	volume = {35},
	doi = {10.1109/TKDE.2023.3236423},
	number = {11},
	journal = {IEEE Trans. Knowl. Data Eng.},
	author = {Han, Jindong and Liu, Hao and Zhu, Hengshu and Xiong, Hui},
	year = {2023},
	pages = {11515--11528},
}

@inproceedings{hu_towards_2024,
	address = {Atlanta GA USA},
	title = {Towards Unifying Diffusion Models for Probabilistic Spatio-Temporal Graph Learning},
	doi = {10.1145/3678717.3691235},
	booktitle = {Proceedings of the 32nd {ACM} {International} {Conference} on {Advances} in {Geographic} {Information} {Systems}},
	publisher = {ACM},
	author = {Hu, Junfeng and Liu, Xu and Fan, Zhencheng and Liang, Yuxuan and Zimmermann, Roger},
	year = {2024},
	pages = {135--146},
}

@article{ji_stden_2022,
	title = {{STDEN}: Towards Physics-Guided Neural Networks for Traffic Flow Prediction},
	volume = {36},
	doi = {10.1609/aaai.v36i4.20322},
	number = {4},
	journal = {Proceedings of the AAAI Conference on Artificial Intelligence},
	author = {Ji, Jiahao and Wang, Jingyuan and Jiang, Zhe and Jiang, Jiawei and Zhang, Hu},
	year = {2022},
	pages = {4048--4056},
}

@article{jin_multivariate_2021,
	title = {Multivariate Air Quality Forecasting With Nested Long Short Term Memory Neural Network},
	volume = {17},
	doi = {10.1109/TII.2021.3065425},
	number = {12},
	journal = {IEEE Trans. Ind. Inf.},
	author = {Jin, Ning and Zeng, Yongkang and Yan, Ke and Ji, Zhiwei},
	year = {2021},
	pages = {8514--8522},
}

@article{liang_airformer_2023,
	title = {{AirFormer}: Predicting Nationwide Air Quality in {China} with Transformers},
	volume = {37},
	doi = {10.1609/aaai.v37i12.26676},
	number = {12},
	journal = {Proceedings of the AAAI Conference on Artificial Intelligence},
	author = {Liang, Yuxuan and Xia, Yutong and Ke, Songyu and Wang, Yiwei and Wen, Qingsong and Zhang, Junbo and Zheng, Yu and Zimmermann, Roger},
	year = {2023},
	pages = {14329--14337},
}

@inproceedings{liu_msdr_2022,
	address = {Washington DC USA},
	title = {{MSDR}: Multi-Step Dependency Relation Networks for Spatial Temporal Forecasting},
	doi = {10.1145/3534678.3539397},
	booktitle = {Proceedings of the 28th {ACM} {SIGKDD} {Conference} on {Knowledge} {Discovery} and {Data} {Mining}},
	publisher = {ACM},
	author = {Liu, Dachuan and Wang, Jin and Shang, Shuo and Han, Peng},
	year = {2022},
	pages = {1042--1050},
}

@inproceedings{liu_residual_2024,
	address = {Seattle, WA, USA},
	title = {Residual Denoising Diffusion Models},
	doi = {10.1109/CVPR52733.2024.00268},
	booktitle = {2024 {IEEE}/{CVF} {Conference} on {Computer} {Vision} and {Pattern} {Recognition} ({CVPR})},
	publisher = {IEEE},
	author = {Liu, Jiawei and Wang, Qiang and Fan, Huijie and Wang, Yinong and Tang, Yandong and Qu, Liangqiong},
	year = {2024},
	pages = {2773--2783},
}

@article{qi_hybrid_2019,
	title = {A hybrid model for spatiotemporal forecasting of {PM2.5} based on graph convolutional neural network and long short-term memory},
	volume = {664},
	doi = {10.1016/j.scitotenv.2019.01.333},
	journal = {Science of The Total Environment},
	author = {Qi, Yanlin and Li, Qi and Karimian, Hamed and Liu, Di},
	year = {2019},
	pages = {1--10},
}

@article{salinas_deepar_2020,
	title = {{DeepAR}: Probabilistic forecasting with autoregressive recurrent networks},
	volume = {36},
	doi = {10.1016/j.ijforecast.2019.07.001},
	number = {3},
	journal = {International Journal of Forecasting},
	author = {Salinas, David and Flunkert, Valentin and Gasthaus, Jan and Januschowski, Tim},
	year = {2020},
	pages = {1181--1191},
}

@article{shuman_emerging_2013,
	title = {The emerging field of signal processing on graphs: Extending high-dimensional data analysis to networks and other irregular domains},
	volume = {30},
	doi = {10.1109/MSP.2012.2235192},
	number = {3},
	journal = {IEEE Signal Process. Mag.},
	author = {Shuman, David I. and Narang, Sunil K. and Frossard, Pascal and Ortega, Antonio and Vandergheynst, Pierre},
	year = {2013},
	pages = {83--98},
}

@article{tao_air_2019,
	title = {Air Pollution Forecasting Using a Deep Learning Model Based on 1D Convnets and Bidirectional GRU},
	volume = {7},
	doi = {10.1109/ACCESS.2019.2921578},
	journal = {IEEE Access},
	author = {Tao, Qing and Liu, Fang and Li, Yong and Sidorov, Denis},
	year = {2019},
	pages = {76690--76698},
}

@inproceedings{wu_quantifying_2021,
	address = {Virtual Event Singapore},
	title = {Quantifying Uncertainty in Deep Spatiotemporal Forecasting},
	doi = {10.1145/3447548.3467325},
	booktitle = {Proceedings of the 27th {ACM} {SIGKDD} {Conference} on {Knowledge} {Discovery} \& {Data} {Mining}},
	publisher = {ACM},
	author = {Wu, Dongxia and Gao, Liyao and Chinazzi, Matteo and Xiong, Xinyue and Vespignani, Alessandro and Ma, Yi-An and Yu, Rose},
	year = {2021},
	pages = {1841--1851},
}

@article{xiao_dual-path_2022,
	title = {A dual-path dynamic directed graph convolutional network for air quality prediction},
	volume = {827},
	doi = {10.1016/j.scitotenv.2022.154298},
	journal = {Science of The Total Environment},
	author = {Xiao, Xiao and Jin, Zhiling and Wang, Shuo and Xu, Jing and Peng, Ziyan and Wang, Rui and Shao, Wei and Hui, Yilong},
	year = {2022},
	pages = {154298},
}

@article{zhang_hybrid_2021,
	title = {A hybrid deep learning technology for {PM2.5} air quality forecasting},
	volume = {28},
	doi = {10.1007/s11356-021-12657-8},
	number = {29},
	journal = {Environ Sci Pollut Res},
	author = {Zhang, Zhendong and Zeng, Yongkang and Yan, Ke},
	year = {2021},
	pages = {39409--39422},
}

@article{zhang_systematic_2024-1,
	title = {A systematic survey of air quality prediction based on deep learning},
	volume = {93},
	doi = {10.1016/j.aej.2024.03.031},
	journal = {Alexandria Engineering Journal},
	author = {Zhang, Zhen and Zhang, Shiqing and Chen, Caimei and Yuan, Jiwei},
	year = {2024},
	pages = {128--141},
}

@article{zheng_gman_2020,
	title = {{GMAN}: A Graph Multi-Attention Network for Traffic Prediction},
	volume = {34},
	doi = {10.1609/aaai.v34i01.5477},
	number = {01},
	journal = {Proceedings of the AAAI Conference on Artificial Intelligence},
	author = {Zheng, Chuanpan and Fan, Xiaoliang and Wang, Cheng and Qi, Jianzhong},
	year = {2020},
	pages = {1234--1241},
}

@article{zhou_theory-guided_2022,
	title = {A theory-guided graph networks based {PM2.5} forecasting method},
	volume = {293},
	doi = {10.1016/j.envpol.2021.118569},
	journal = {Environmental Pollution},
	author = {Zhou, Hongye and Zhang, Feng and Du, Zhenhong and Liu, Renyi},
	year = {2022},
	pages = {118569},
}

@inproceedings{cachay_dyffusion_2023,
	title = {{DYffusion}: A Dynamics-informed Diffusion Model for Spatiotemporal Forecasting},
	volume = {36},
	url = {https://proceedings.neurips.cc/paper_files/paper/2023/hash/8df90a1440ce782d1f5607b7a38f2531-Abstract-Conference.html},
	booktitle = {Advances in Neural Information Processing Systems},
	publisher = {Curran Associates, Inc.},
	author = {R{\"u}hling Cachay, Salva and Zhao, Bo and Joren, Hailey and Yu, Rose},
	year = {2023},
	pages = {45259--45287},
}

@inproceedings{chen_neural_2019,
	title = {Neural Ordinary Differential Equations},
	volume = {31},
	url = {https://proceedings.neurips.cc/paper_files/paper/2018/hash/69386f6bb1dfed68692a24c8686939b9-Abstract.html},
	booktitle = {Advances in {Neural} {Information} {Processing} {Systems}},
	publisher = {Curran Associates, Inc.},
	author = {Chen, Ricky T. Q. and Rubanova, Yulia and Bettencourt, Jesse and Duvenaud, David K.},
	year = {2018},
}

@article{guo_attention_2019,
	title = {Attention Based Spatial-Temporal Graph Convolutional Networks for Traffic Flow Forecasting},
	volume = {33},
	doi = {10.1609/aaai.v33i01.3301922},
	number = {01},
	journal = {Proceedings of the AAAI Conference on Artificial Intelligence},
	author = {Guo, Shengnan and Lin, Youfang and Feng, Ning and Song, Chao and Wan, Huaiyu},
	year = {2019},
	pages = {922--929},
}

@inproceedings{hettige_airphynet_2024,
	title = {{AirPhyNet}: Harnessing Physics-Guided Neural Networks for Air Quality Prediction},
	booktitle = {International Conference on Learning Representations},
	url = {https://openreview.net/forum?id=JW3jTjaaAB},
	author = {Hettige, Kethmi Hirushini and Ji, Jiahao and Xiang, Shili and Long, Cheng and Cong, Gao and Wang, Jingyuan},
	year = {2024},
}

@inproceedings{ho_denoising_2020,
	title = {Denoising Diffusion Probabilistic Models},
	volume = {33},
	url = {https://proceedings.neurips.cc/paper/2020/hash/4c5bcfec8584af0d967f1ab10179ca4b-Abstract.html},
	booktitle = {Advances in {Neural} {Information} {Processing} {Systems}},
	publisher = {Curran Associates, Inc.},
	author = {Ho, Jonathan and Jain, Ajay and Abbeel, Pieter},
	year = {2020},
	pages = {6840--6851},
}

@inproceedings{li_diffusion_2018,
	title = {Diffusion Convolutional Recurrent Neural Network: Data-Driven Traffic Forecasting},
	booktitle = {International Conference on Learning Representations},
	url = {https://openreview.net/forum?id=SJiHXGWAZ},
	author = {Li, Yaguang and Yu, Rose and Shahabi, Cyrus and Liu, Yan},
	year = {2018},
}

@inproceedings{lin_specstg_2024,
	title = {{SpecSTG}: A Fast Spectral Diffusion Framework for Probabilistic Spatio-Temporal Traffic Forecasting},
	doi = {10.1109/IJCNN64981.2025.11227295},
	booktitle = {2025 {International} {Joint} {Conference} on {Neural} {Networks} ({IJCNN})},
	author = {Lin, Lequan and Shi, Dai and Han, Andi and Gao, Junbin},
	publisher = {IEEE},
	year = {2025},
	pages = {1--10},
}

@article{matheson_scoring_1976,
	title = {Scoring Rules for Continuous Probability Distributions},
	volume = {22},
	url = {https://www.jstor.org/stable/2629907},
	number = {10},
	journal = {Management Science},
	publisher = {INFORMS},
	author = {Matheson, James E. and Winkler, Robert L.},
	year = {1976},
	pages = {1087--1096},
}

@article{raissi_physics_2017,
	title = {Physics-informed neural networks: A deep learning framework for solving forward and inverse problems involving nonlinear partial differential equations},
	volume = {378},
	doi = {10.1016/j.jcp.2018.10.045},
	journal = {Journal of Computational Physics},
	author = {Raissi, Maziar and Perdikaris, Paris and Karniadakis, George Em},
	year = {2019},
	pages = {686--707},
}

@inproceedings{rubanova_latent_2019,
	title = {Latent Ordinary Differential Equations for Irregularly-Sampled Time Series},
	volume = {32},
	url = {https://proceedings.neurips.cc/paper_files/paper/2019/hash/42a6845a557bef704ad8ac9cb4461d43-Abstract.html},
	booktitle = {Advances in {Neural} {Information} {Processing} {Systems}},
	publisher = {Curran Associates, Inc.},
	author = {Rubanova, Yulia and Chen, Ricky T. Q. and Duvenaud, David K.},
	year = {2019},
}

@inproceedings{shi_resfusion_2024,
	title = {{Resfusion}: Denoising Diffusion Probabilistic Models for Image Restoration Based on Prior Residual Noise},
	volume = {37},
	doi = {10.52202/079017-4153},
	booktitle = {Advances in Neural Information Processing Systems},
	publisher = {Curran Associates, Inc.},
	author = {Shi, Zhenning and Zheng, Haoshuai and Xu, Chen and Dong, Changsheng and Pan, Bin and Xie, Xueshuo and He, Along and Li, Tao and Fu, Huazhu},
	year = {2024},
	pages = {130664--130693},
}

@inproceedings{tashiro_csdi_2021,
	title = {{CSDI}: Conditional Score-based Diffusion Models for Probabilistic Time Series Imputation},
	volume = {34},
	url = {https://papers.neurips.cc/paper_files/paper/2021/hash/cfe8504bda37b575c70ee1a8276f3486-Abstract.html},
	booktitle = {Advances in {Neural} {Information} {Processing} {Systems}},
	publisher = {Curran Associates, Inc.},
	author = {Tashiro, Yusuke and Song, Jiaming and Song, Yang and Ermon, Stefano},
	year = {2021},
	pages = {24804--24816},
}

@inproceedings{yu_spatio-temporal_2018,
	title = {Spatio-Temporal Graph Convolutional Networks: A Deep Learning Framework for Traffic Forecasting},
	booktitle = {Proceedings of the Twenty-Seventh International Joint Conference on Artificial Intelligence ({IJCAI})},
	url = {https://www.ijcai.org/proceedings/2018/505},
	doi = {10.24963/ijcai.2018/505},
	author = {Yu, Bing and Yin, Haoteng and Zhu, Zhanxing},
	year = {2018},
	pages = {3634--3640},
}

@inproceedings{yue_resshift_2023,
	title = {{ResShift}: Efficient Diffusion Model for Image Super-resolution by Residual Shifting},
	volume = {36},
	url = {https://proceedings.neurips.cc/paper_files/paper/2023/hash/2ac2eac5098dba08208807b65c5851cc-Abstract-Conference.html},
	booktitle = {Advances in Neural Information Processing Systems},
	publisher = {Curran Associates, Inc.},
	author = {Yue, Zongsheng and Wang, Jianyi and Loy, Chen Change},
	year = {2023},
	pages = {13294--13307},
}

@inproceedings{wang_pm25-gnn_2021,
	title = {{PM2.5-GNN}: A Domain Knowledge Enhanced Graph Neural Network For {PM2.5} Forecasting},
	doi = {10.1145/3397536.3422208},
	booktitle = {Proceedings of the 28th ACM International Conference on Advances in Geographic Information Systems},
	author = {Wang, Shuo and Li, Yanran and Zhang, Jiang and Meng, Qingye and Meng, Lingwei and Gao, Fei},
	publisher = {ACM},
	year = {2020},
	pages = {163--166},
}

@inproceedings{wen_diffstg_2023,
	title = {{DiffSTG}: Probabilistic Spatio-Temporal Graph Forecasting with Denoising Diffusion Models},
	doi = {10.1145/3589132.3625614},
	booktitle = {Proceedings of the 31st {ACM} International Conference on Advances in Geographic Information Systems},
	author = {Wen, Haomin and Lin, Youfang and Xia, Yutong and Wan, Huaiyu and Wen, Qingsong and Zimmermann, Roger and Liang, Yuxuan},
	publisher = {ACM},
	year = {2023},
	pages = {1--12},
}

@article{zeng_are_2022,
	title = {Are Transformers Effective for Time Series Forecasting?},
	journal = {Proceedings of the AAAI Conference on Artificial Intelligence},
	author = {Zeng, Ailing and Chen, Muxi and Zhang, Lei and Xu, Qiang},
	year = {2023},
	volume = {37},
	number = {9},
	pages = {11121--11128},
	doi = {10.1609/aaai.v37i9.26317},
}

@inproceedings{tian_air_2024,
	title = {Air Quality Prediction with Physics-Guided Dual Neural {ODEs} in Open Systems},
	booktitle = {International Conference on Learning Representations},
	url = {https://openreview.net/forum?id=kOJf7Dklyv},
	author = {Tian, Jindong and Liang, Yuxuan and Xu, Ronghui and Chen, Peng and Guo, Chenjuan and Zhou, Aoying and Pan, Lujia and Rao, Zhongwen and Yang, Bin},
	year = {2025},
}

@inproceedings{LiuMingzhe2023PACD,
  title = {{{PriSTI}}: {{A Conditional Diffusion Framework}} for {{Spatiotemporal Imputation}}},
  author = {Liu, Mingzhe and Huang, Han and Feng, Hao and Sun, Leilei and Du, Bowen and Fu, Yanjie},
  booktitle = {2023 IEEE 39th International Conference on Data Engineering (ICDE)},
  publisher = {IEEE},
  year = {2023},
  pages = {1927--1939},
  doi = {10.1109/ICDE55515.2023.00150}
}

@article{wu_autocts_2022,
  title = {{{AutoCTS}}: Automated Correlated Time Series Forecasting},
  author = {Wu, Xinle and Zhang, Dalin and Guo, Chenjuan and He, Chaoyang and Yang, Bin and Jensen, Christian S.},
  year = {2022},
  journal = {Proceedings of the VLDB Endowment},
  volume = {15},
  number = {4},
  pages = {971--983},
  issn = {2150-8097},
  doi = {10.14778/3503585.3503604},
  url = {https://dl.acm.org/doi/10.14778/3503585.3503604}
}

@inproceedings{song_denoising_2021,
  title = {Denoising Diffusion Implicit Models},
  author = {Song, Jiaming and Meng, Chenlin and Ermon, Stefano},
  booktitle = {International Conference on Learning Representations},
  year = {2021},
  url = {https://openreview.net/forum?id=St1giarCHLP}
}

@inproceedings{gal_dropout_2016,
  title = {Dropout as a Bayesian Approximation: Representing Model Uncertainty in Deep Learning},
  author = {Gal, Yarin and Ghahramani, Zoubin},
  booktitle = {Proceedings of the 33rd International Conference on Machine Learning},
  series = {PMLR},
  volume = {48},
  pages = {1050--1059},
  year = {2016},
  url = {https://proceedings.mlr.press/v48/gal16.html}
}

@inproceedings{stankeviciute_conformal_2021,
  title = {Conformal Time-series Forecasting},
  author = {Stankevi{\v{c}}i{\=u}t{\.e}, Kamil{\.e} and Alaa, Ahmed M. and {van der Schaar}, Mihaela},
  booktitle = {Advances in Neural Information Processing Systems},
  volume = {34},
  publisher = {Curran Associates, Inc.},
  pages = {6216--6228},
  year = {2021},
  url = {https://proceedings.neurips.cc/paper/2021/hash/312f1ba2a72318edaaa995a67835fad5-Abstract.html}
}

@book{vovk_algorithmic_2005,
  doi = {10.1007/b106715},
  title = {Algorithmic Learning in a Random World},
  author = {Vovk, Vladimir and Gammerman, Alexander and Shafer, Glenn},
  publisher = {Springer},
  year = {2005}
}

@inproceedings{rasul_autoregressive_2021,
  title = {Autoregressive Denoising Diffusion Models for Multivariate Probabilistic Time Series Forecasting},
  author = {Rasul, Kashif and Seward, Calvin and Schuster, Ingmar and Vollgraf, Roland},
  booktitle = {Proceedings of the 38th International Conference on Machine Learning},
  series = {PMLR},
  volume = {139},
  pages = {8857--8868},
  year = {2021},
  url = {https://proceedings.mlr.press/v139/rasul21a.html}
}

@inproceedings{li_transformer_2024,
  title = {Transformer-Modulated Diffusion Models for Probabilistic Multivariate Time Series Forecasting},
  author = {Li, Yuxin and Chen, Wenchao and Hu, Xinyue and Chen, Bo and Sun, Baolin and Zhou, Mingyuan},
  booktitle = {International Conference on Learning Representations},
  year = {2024},
  url = {https://openreview.net/forum?id=qae04YACHs}
}

@inproceedings{kondor_diffusion_2002,
  title = {Diffusion Kernels on Graphs and Other Discrete Input Spaces},
  author = {Kondor, Risi Imre and Lafferty, John D.},
  booktitle = {Proceedings of the Nineteenth International Conference on Machine Learning},
  publisher = {Morgan Kaufmann},
  pages = {315--322},
  year = {2002}
}

@inproceedings{erdm_2025,
  doi = {10.52202/085713-0068},
  title = {Elucidated Rolling Diffusion Models for Probabilistic Forecasting of Complex Dynamics},
  author = {R{\"u}hling Cachay, Salva and Aittala, Miika and Kreis, Karsten and Brenowitz, Noah and Vahdat, Arash and Mardani, Morteza and Yu, Rose},
  booktitle = {Advances in Neural Information Processing Systems},
  volume = {38},
  pages = {2098--2137},
  year = {2025}
}

@article{mardani_corrdiff_2025,
  doi = {10.1038/s43247-025-02042-5},
  articleno = {124},
  title = {Residual Corrective Diffusion Modeling for {km}-scale Atmospheric Downscaling},
  author = {Mardani, Morteza and Brenowitz, Noah and Cohen, Yair and Pathak, Jaideep and Chen, Chieh-Yu and Liu, Cheng-Chin and Vahdat, Arash and Nabian, Mohammad Amin and Ge, Tao and Subramaniam, Akshay and Kashinath, Karthik and Kautz, Jan and Pritchard, Mike},
  journal = {Communications Earth \& Environment},
  volume = {6},
  year = {2025},
}

@misc{ntmaehara_lowpass_2019,
  title = {Revisiting Graph Neural Networks: All We Have is Low-Pass Filters},
  author = {{NT}, Hoang and Maehara, Takanori},
  howpublished = {arXiv:1905.09550},
  doi = {10.48550/arXiv.1905.09550},
  year = {2019}
}

@article{bo_beyond_2021,
  title = {Beyond Low-frequency Information in Graph Convolutional Networks},
  author = {Bo, Deyu and Wang, Xiao and Shi, Chuan and Shen, Huawei},
  journal = {Proceedings of the AAAI Conference on Artificial Intelligence},
  volume = {35},
  number = {5},
  pages = {3950--3957},
  doi = {10.1609/aaai.v35i5.16514},
  year = {2021}
}

@inproceedings{wu_sgc_2019,
  url = {https://proceedings.mlr.press/v97/wu19e.html},
  title = {Simplifying Graph Convolutional Networks},
  author = {Wu, Felix and Souza, Amauri H. and Zhang, Tianyi and Fifty, Christopher and Yu, Tao and Weinberger, Kilian Q.},
  booktitle = {Proceedings of the 36th International Conference on Machine Learning},
  series = {PMLR},
  volume = {97},
  pages = {6861--6871},
  year = {2019}
}

@inproceedings{defferrard_chebnet_2016,
  title = {Convolutional Neural Networks on Graphs with Fast Localized Spectral Filtering},
  author = {Defferrard, Micha\"{e}l and Bresson, Xavier and Vandergheynst, Pierre},
  booktitle = {Advances in Neural Information Processing Systems},
  volume = {29},
  publisher = {Curran Associates, Inc.},
  pages = {3844--3852},
  year = {2016},
  url = {https://papers.nips.cc/paper/6081-convolutional-neural-networks-on-graphs-with-fast-localized-spectral-filtering}
}

@inproceedings{chien_gprgnn_2021,
  title = {Adaptive Universal Generalized {PageRank} Graph Neural Network},
  author = {Chien, Eli and Peng, Jianhao and Li, Pan and Milenkovic, Olgica},
  booktitle = {International Conference on Learning Representations},
  year = {2021},
  url = {https://openreview.net/forum?id=n6jl7fLxrP}
}

@inproceedings{liu_gmlp_2021,
  title = {Pay Attention to {MLPs}},
  author = {Liu, Hanxiao and Dai, Zihang and So, David R. and Le, Quoc V.},
  booktitle = {Advances in Neural Information Processing Systems},
  volume = {34},
  publisher = {Curran Associates, Inc.},
  pages = {9204--9215},
  year = {2021},
  url = {https://papers.nips.cc/paper/2021/hash/4cc05b35c2f937c5bd9e7d41d3686fff-Abstract.html}
}

@inproceedings{jo_gdss_2022,
  title = {Score-based Generative Modeling of Graphs via the System of Stochastic Differential Equations},
  author = {Jo, Jaehyeong and Lee, Seul and Hwang, Sung Ju},
  booktitle = {Proceedings of the 39th International Conference on Machine Learning},
  series = {PMLR},
  volume = {162},
  pages = {10362--10383},
  year = {2022},
  url = {https://proceedings.mlr.press/v162/jo22a.html}
}

\newpage
\appendix
\section*{Appendix}

\section{DD against CSDI: seed statistics and significance}
\label{app:seeds}

Every row of Table~\ref{tab:main} is a seed mean; this appendix reports the DD-against-CSDI comparison at full resolution because it decides the headline CRPS claim. Table~\ref{tab:seeds} gives the seed means with sample standard deviations, and Table~\ref{tab:sig} the significance tests behind Section~\ref{sec:results}; the per-seed values are in the code repository rather than here, for space.

\begin{table}[h]
\caption{DD against CSDI, three seeds each, mean $\pm$ sample standard deviation; bold marks the lower mean. The comparison is unpaired.}
\label{tab:seeds}
\footnotesize\setlength{\tabcolsep}{3pt}
\begin{tabular}{llrrr}
\toprule
Dataset & Method & MAE & RMSE & CRPS \\
\midrule
\multicolumn{5}{l}{\emph{Initial}} \\
Beijing & CSDI & $20.98 \pm 0.79$ & $\mathbf{52.51 \pm 1.04}$ & $0.3548 \pm 0.0150$ \\
        & DD   & $\mathbf{20.68 \pm 0.75}$ & $54.29 \pm 2.02$ & $\mathbf{0.3206 \pm 0.0158}$ \\
Athens  & CSDI & $\mathbf{8.65 \pm 0.15}$ & $\mathbf{15.97 \pm 0.32}$ & $0.3551 \pm 0.0080$ \\
        & DD   & $8.90 \pm 0.11$ & $16.12 \pm 0.30$ & $\mathbf{0.3273 \pm 0.0060}$ \\
PEMS08  & CSDI & $\mathbf{5.80 \pm 0.06}$ & $\mathbf{14.91 \pm 0.16}$ & $\mathbf{0.1247 \pm 0.0011}$ \\
        & DD   & $6.16 \pm 0.17$ & $15.63 \pm 0.27$ & $0.1282 \pm 0.0024$ \\
PEMS04  & CSDI & $\mathbf{7.69 \pm 0.10}$ & $\mathbf{19.50 \pm 0.28}$ & $0.1380 \pm 0.0020$ \\
        & DD   & $7.97 \pm 0.01$ & $20.42 \pm 0.07$ & $\mathbf{0.1336 \pm 0.0001}$ \\
\midrule
\multicolumn{5}{l}{\emph{Rolling}} \\
Beijing & CSDI & $19.76 \pm 0.77$ & $\mathbf{50.87 \pm 1.04}$ & $0.3351 \pm 0.0155$ \\
        & DD   & $\mathbf{19.66 \pm 0.91}$ & $54.11 \pm 2.02$ & $\mathbf{0.3032 \pm 0.0171}$ \\
Athens  & CSDI & $8.30 \pm 0.17$ & $15.59 \pm 0.33$ & $0.3396 \pm 0.0084$ \\
        & DD   & $\mathbf{8.22 \pm 0.10}$ & $\mathbf{15.24 \pm 0.24}$ & $\mathbf{0.3016 \pm 0.0047}$ \\
PEMS08  & CSDI & $\mathbf{5.58 \pm 0.05}$ & $\mathbf{14.33 \pm 0.16}$ & $\mathbf{0.1184 \pm 0.0009}$ \\
        & DD   & $5.78 \pm 0.15$ & $14.83 \pm 0.25$ & $0.1207 \pm 0.0020$ \\
PEMS04  & CSDI & $\mathbf{7.40 \pm 0.07}$ & $\mathbf{18.81 \pm 0.36}$ & $0.1308 \pm 0.0014$ \\
        & DD   & $7.42 \pm 0.01$ & $19.21 \pm 0.08$ & $\mathbf{0.1245 \pm 0.0002}$ \\
\bottomrule
\end{tabular}
\end{table}

Table~\ref{tab:sig} reports the Welch two-sample tests on the three initial-view CRPS values per side, with the Holm correction across the four datasets and approximate $95\%$ Welch intervals for the CRPS difference, DD minus CSDI, so that negative favors DD. Only Athens clears a conventional threshold and survives the correction. The test is unpaired, and a non-rejection on PEMS08 is not evidence of equivalence, so the summary on CRPS is that DD is significantly better on one dataset, better but not significantly on two, and worse but not significantly on one. Every test is on the initial view, and the intervals describe training-seed variation under this protocol rather than deployment uncertainty.

\begin{table}[h]
\caption{Initial-view CRPS comparison of DD and CSDI (three seeds per method): Welch $p$-values, Holm-adjusted values over four datasets, and approximate $95\%$ intervals for DD minus CSDI; negative differences favor DD.}
\label{tab:sig}
\footnotesize\setlength{\tabcolsep}{4pt}
\begin{tabular}{lrrr}
\toprule
Dataset & $p$ & Holm & $95\%$ CI on $\Delta$CRPS \\
\midrule
Beijing & 0.053 & 0.159 & $-0.034\,[-0.069, +0.001]$ \\
Athens & 0.011 & 0.042 & $-0.028\,[-0.044, -0.011]$ \\
PEMS08 & 0.113 & 0.159 & $+0.004\,[-0.002, +0.009]$ \\
PEMS04 & 0.061 & 0.159 & $-0.004\,[-0.009, +0.001]$ \\
\bottomrule
\end{tabular}
\end{table}

\section{The closed form for the start step}
\label{app:sprime}

\begin{table}[t]
\caption{Beijing sampling time (ms per window) against scheduled chain length $S$, with the fitted $b$ in $T \propto S^{b}$. Column heads give $S$ with DD's executed $S'$ in parentheses; the full-chain baselines execute all $S$ steps and DDIM25 always 25. Cells follow the sweep protocol of Section~\ref{sec:latency} (batch size 16) and are not comparable with the operating points of Table~\ref{tab:latency}.}
\label{tab:scaling}
\footnotesize\setlength{\tabcolsep}{2.4pt}
\begin{tabular}{lrrrrrr}
\toprule
Model & $100\,(52)$ & $200\,(74)$ & $300\,(91)$ & $400\,(105)$ & $500\,(117)$ & $b$ \\
\midrule
DD & 856 & 1221 & 1488 & 1732 & 1910 & \textbf{0.501} \\
DiffSTG & 1029 & 2057 & 3082 & 4110 & 5140 & 0.999 \\
CSDI & 3281 & 6560 & 9836 & 13110 & 16488 & 1.002 \\
DiffSTG (DDIM25) & 255 & 255 & 255 & 255 & 255 & 0.000 \\
\bottomrule
\end{tabular}
\end{table}

Equation~\ref{eq:sprimeclosed} is our derivation rather than Resfusion's, which computes $S'$ by the discrete minimization of Equation~\ref{eq:sprime}, so we record it here. For the linear schedule, $\beta_s \approx (s/S)\,\beta_S$, since the shipped $\beta_1{=}10^{-4}$ is negligible, and two elementary approximations then give the closed form. First, $\log \bar\alpha_s = \sum_{i=1}^{s} \log(1-\beta_i) \approx -\sum_{i=1}^{s} \beta_i$, valid while every $\beta_i \ll 1$; the leading neglected term is $\sum_{i} \beta_i^2/2$. Second, $\sum_{i=1}^{s} \beta_i \approx \beta_S\,s(s{+}1)/2S \approx \beta_S s^2/2S$, dropping a relative $1/s$. Together $\sqrt{\bar\alpha_s} \approx \exp(-\beta_S s^2/4S)$, and setting this to one half yields Equation~\ref{eq:sprimeclosed}, which depends on the schedule only through $S/\beta_S$ and grows as $\sqrt{S}$ at fixed $\beta_S$. Both neglected corrections push $\sqrt{\bar\alpha_s}$ slightly below the leading estimate; the shipped configuration sits at $\sqrt{\bar\alpha_{S'}}{=}0.505$ rather than exactly one half chiefly because the discrete minimizer sits just below the real-valued crossing. At $(S,\beta_S){=}(100,0.1)$ the closed form gives $52.7$ against the exact discrete minimizer $52$, and across $S \in \{100,\dots,500\}$ it matches the executed step counts of Table~\ref{tab:scaling}, the raw Beijing timings behind Figure~\ref{fig:scaling}, to under one step. Under the leading-order approximation $S'$ depends only on $S/\beta_S$; the exact integer minimizer does not inherit this property in general, since the nonzero $\beta_1$, the endpoint indexing, and the integer minimization all enter (for example, $(400,\,0.4)$ gives $53$ against $52$ for $(100,\,0.1)$), though the two equal-ratio pairs tested in Section~\ref{sec:ablation} retain identical exact step counts.

\section{DDIM acceleration of DiffSTG}
\label{app:ddim}

DDIM sampling at $25$ steps reuses the trained DiffSTG checkpoints unchanged and swaps in a strided sampler on a uniform $25$-index grid over the $S{=}200$ schedule, with no retraining; it is by far the cheapest learned sampler in Table~\ref{tab:latency}, and the speed is bought with accuracy (Table~\ref{tab:ddim}). The sampler's stochasticity parameter $\eta$ drives an even sharper sensitivity. The DiffSTG reference implementation fixes $\eta{=}1$. The DDIM25 rows of Table~\ref{tab:main} follow it. Table~\ref{tab:ddim} pairs every $\eta{=}1$ cell against its own DDPM parent checkpoint at matched seeds: on the air-quality networks the acceleration costs at most $1.1\%$ initial-view MAE and improves one Athens seed, while on the traffic networks it costs $7$ to $20\%$.

The deterministic $\eta{=}0$ variant of the same sampler amplifies the traffic cost sharply, $41.8\%$ CRPS on PEMS08 against $0.3$ to $2.2\%$ CRPS on the air-quality networks, a sensitivity to the sampler's stochasticity at a $25$-step stride whose mechanism we leave open. The timing in Table~\ref{tab:latency} was measured at the $\eta{=}0$ configuration and is reported as such; the executed step count is identical at either $\eta$.

\begin{table}[!b]
\caption{DiffSTG initial-view MAE for DDPM and 25-step DDIM ($\eta{=}1$) from the same checkpoints at seeds 0--2 ($S{=}200$, $\beta_S{=}0.1$). Percent changes use unrounded values.}
\label{tab:ddim}
\footnotesize\setlength{\tabcolsep}{4pt}
\begin{tabular}{lrrr}
\toprule
Dataset & DDPM & DDIM25 ($\eta{=}1$) & Change (\%) \\
\midrule
Beijing & 22.29 / 21.00 / 21.76 & 22.38 / 21.13 / 21.99 & $+0.4$ / $+0.6$ / $+1.1$ \\
Athens & 10.52 / 11.07 / 11.58 & 10.63 / 11.16 / 11.46 & $+1.0$ / $+0.8$ / $-1.1$ \\
PEMS08 & 9.53 / 7.26 / 9.44 & 10.65 / 7.76 / 10.96 & $+11.7$ / $+6.8$ / $+16.1$ \\
PEMS04 & 13.44 / 16.33 / 15.34 & 15.56 / 18.73 / 18.43 & $+15.8$ / $+14.7$ / $+20.2$ \\
\bottomrule
\end{tabular}
\end{table}

\end{document}